# Beyond SBDD: Geometric Deep Learning in Polypharmacology and Multi-target Drug Design

Tianming Han[1,2,#], Zhijie Pan[1,#], Wenchi Ge[1,#], Qi Zhao[1,*]

[1]School of Computer Science and Software Engineering, University of Science and Technology Liaoning, Anshan, 114051, China

[2]EFREI Engineering School of Digital Technologies, Paris-Panthéon-Assas University, Villejuif, 94800, France

[#]These authors contributed equally to the paper as first authors.

*Corresponding author

**Email:** zhaoqi@lnu.edu.cn

## Abstract:

The traditional "one drug, one target" paradigm of structure-based drug design (SBDD) frequently proves inadequate for treating multifactorial diseases such as cancer and neurodegenerative disorders, owing to compensatory signaling pathways and the emergence of drug resistance. While polypharmacology offers a synergistic therapeutic strategy, the rational design of ligands capable of simultaneously satisfying the geometric constraints imposed by multiple targets remains a major computational bottleneck. This review positions geometric deep learning (GDL) as a powerful integrative approach to overcome these limitations. We systematically survey GDL architectures ranging from invariant graph neural networks to SE(3)-equivariant diffusion models that harness non-Euclidean molecular data to capture intrinsic three-dimensional (3D) structural interdependencies. We critically analyze GDL applications across three core dimensions, including the characterization of shared binding pockets via geometric embeddings, multi-target bioactivity prediction through heterogeneous graph fusion, and *de novo* generation of dual-target ligands. Particular emphasis is placed on emerging structure-conditioned generative algorithms that integrate diffusion models with reinforcement learning to autonomously resolve complex geometric conflicts between competing binding sites. Furthermore, we evaluate the pivotal role of multimodal omics integration and specialized geometric benchmarking infrastructures in validating these models. By synthesizing these methodological advances, this review elucidates the paradigm shift in drug discovery from serendipitous exploration to rational, structure-driven polypharmacological molecular engineering, thereby providing a clear, structured guide for navigating the complexities of next-generation therapeutics.

**Keywords:** geometric deep learning, polypharmacology, multi-target drug design, generative AI, structure-based design

## 1. Introduction

Traditional structure-based drug design (SBDD) has long adhered to the “one drug, one target” paradigm [1], positioning a single target as the focal point of drug discovery. This precision-targeted “magic bullet” strategy achieves notable success against simple diseases but falls short against complex pathologies such as cancer, metabolic disorders, and neurodegenerative diseases [2]. Single-target inhibition is often circumvented by compensatory signaling pathways, and pathogens or cancer cells readily develop resistance through mutations [3]. Concurrently, clinical management of complex diseases often relies on multidrug combination therapies, which heighten the risks of drug–drug interactions and adverse reactions, complicate dosing regimens, and reduce patient adherence.

Polypharmacology has emerged as a promising therapeutic paradigm for complex diseases and is garnering considerable attention. Unlike the traditional “one drug, one target” approach, the “one drug, multiple targets” strategy aims to design molecules that simultaneously modulate multiple interrelated targets, thereby eliciting synergistic effects at the systems level. This approach can simplify treatment regimens, as a single multi-target agent reduces administration frequency and variety compared with combination therapies, thereby mitigating risks of drug–drug interactions and adverse reactions. Moreover, requiring patients to take only one medication is expected to substantially enhance compliance. Indeed, in oncology, neurology, and infectious diseases, network-directed multi-target interventions have demonstrated distinct advantages [4–6]. Many successful drugs initially dismissed as “dirty drugs” for their promiscuous profiles owe much of their clinical efficacy to this multi-target property. Consequently, drug discovery is shifting from conventional single-target strategies toward networked multi-target interventions that more effectively address biological redundancy and compensatory mechanisms.

Traditional SBDD methods typically address targets in isolation, relying primarily on virtual screening techniques such as molecular docking [7]. These methods employ predefined scoring functions to evaluate static ligand–receptor interactions and perform exhaustive searches in vast chemical spaces. However, this target-centric and static approach faces multiple challenges in multi-target drug design. First, sequential optimization for individual targets may overlook candidates capable of balanced binding across multiple sites. Second, simultaneous combinatorial screening against multiple targets leads to an explosive expansion of the search space and prohibitive computational costs. Third, conventional multi-target optimization often follows sequential and suboptimal modification trajectories, such as first optimizing for the primary target before accommodating secondary targets. Lacking a global 3D synergistic perspective, this stepwise iteration frequently converges to local optima. Consequently, the resulting molecules frequently fail to achieve the desired multi-target binding equilibrium in dynamic physiological environments. Despite SBDD's profound influence on drug development, its toolbox for deliberate multi-target design remains limited, underscoring the urgent need for novel computational paradigms to navigate multi-target drug discovery complexities [8]. The core challenge lies in designing ligands that simultaneously satisfy the binding requirements of two or more proteins, a task that is orders of magnitude more difficult than single-target optimization.

Geometric deep learning (GDL), a machine-learning paradigm tailored for non-Euclidean data such as graphs and molecular surfaces, provides a powerful framework for overcoming the aforementioned limitations. Although two-dimensional (2D) molecular topological graph neural networks constitute a classic GDL application, target recognition fundamentally hinges on precise 3D shape complementarity. Accordingly, GDL methodologies in drug design are undergoing a pivotal shift from “2D topology” toward “3D spatial geometry” [9]. In contrast to models restricted to one-dimensional (1D) sequences or simple 2D connectivity, contemporary 3D GDL frameworks explicitly encode the 3D coordinates and spatial symmetries of both proteins and ligands. This capacity is particularly vital in multi-target design, where a single ligand must simultaneously satisfy the stringent spatial constraints of multiple receptor binding pockets. Studies have shown that incorporating 3D conformational information, rather than relying solely on 2D features, substantially enhances the accuracy of multi-target activity prediction and molecular generation, underscoring the decisive advantage of 3D geometric representations in resolving complex spatial conflicts. Concurrently, landmark advances in structural biology, exemplified by AlphaFold, have vastly expanded the available protein structure space[10–12]. High-quality structural models are now accessible for thousands of human proteins and protein complexes, furnishing a robust foundation for computer-aided design. The integration of large-scale structural data (e.g., predicted proteomes) with geometric AI has laid the groundwork for addressing multi-target design in an unprecedented way, enabling algorithms to autonomously learn the “rules of molecular recognition” across diverse target classes directly in 3D space and obviating the need for manually defined pharmacophores or single-target docking scores.

Preliminary studies confirm that GDL effectively surmounts several key barriers in polypharmacology. GDL models exhibit outstanding performance in detecting and exploiting 3D patterns that connect proteins and ligands across disparate biological systems. For instance, GDL-based pocket-detection tools now identify “druggable” binding sites on proteins with far greater accuracy than conventional geometric algorithms [13,14]. Furthermore, GDL is transforming the characterization and comparative analysis of protein binding sites. Recent latent-space models embed binding pockets as high-dimensional vectors, cluster them by similarity, and even correlate geometric features with ligand preferences [15]. This facilitates rapid target identification, such as aligning allosteric sites of one protein with active sites of another to predict cross-reactivity or facilitate drug repositioning. These tasks were previously difficult to scale.

In ligand design, GDL is propelling multi-target generative approaches far beyond the limits of human intuition or brute-force docking. These GDL-driven methods generate novel ligands that display superior binding affinity and selectivity across multiple targets, often surpassing docking-based strategies in the creation of drug-like analogs [16,17]. In addition to fragment linking, researchers have introduced 3D diffusion models and reinforcement learning frameworks that construct candidate molecules atom-by-atom inside target structures under multi-objective reward guidance [18–20]. These generative models explicitly incorporate 3D shape complementarity with two proteins simultaneously, an achievement impossible with traditional QSAR or 2D methods. Thus, GDL not only predicts polypharmacological activity but also directly “creates” entirely new molecules endowed with polypharmacological properties, facilitating the evolution of multi-target drug design from serendipitous discovery (“magic shotgun”–style dirty drugs) to a rational, AI-driven paradigm.

Despite the rising prominence of multi-target strategies, existing reviews reveal notable gaps in this interdisciplinary domain. Some focus predominantly on network pharmacology or traditional multi-target screening without systematically addressing recent advances in GDL for multi-target generative design and activity prediction, while others omit a comprehensive overview of geometric representation methods for protein binding pockets, thereby hindering readers’ full appreciation of GDL’s potential in multi-target contexts. To address these deficiencies, the present review seeks to delineate a new paradigm for multi-target drug design. Its structure proceeds as follows. First, the fundamental principles and methodological foundations of GDL for modeling molecular and protein structures are introduced. Second, GDL applications in the 3D identification of multi-target binding pockets and high-dimensional latent-space representations are examined. Subsequently, state-of-the-art generative GDL architectures are analyzed in depth, demonstrating how they resolve complex inter-receptor spatial geometric conflicts end-to-end to enable rational customization of multi-target ligands. Next, the scope expands from microscopic structures to macroscopic systems, exploring mechanisms for integrating cross-modal omics data with 3D geometric features while systematically evaluating experimental validation strategies and architectural interpretability of multi-target models. Additionally, this work uniquely synthesizes the data infrastructure underpinning geometric polypharmacology together with principles for constructing standardized benchmarks. Finally, we prospectively discuss the core challenges confronting the field, including surmounting the static rigid-body assumption, achieving multi-dimensional Pareto optimality of properties, and establishing automated "dry–wet experimental closed loops". Through this systematic review, we provide a comprehensive reference framework and roadmap for next-generation multi-target drug design as illustrated in Figure 1, thereby accelerating the maturation of an innovative drug-discovery paradigm that transcends conventional single-target approaches.

# 2. GDL algorithms and its applications in multi-target drug design

## 2.1 GDL for multi-target drug modeling: background and requirements

The clinical success of the multi-target strategy has been convincingly demonstrated by several approved drugs. Representative examples include the classic antipsychotic clozapine [21], widely regarded as the "Magic Shotgun" of polypharmacology. By simultaneously antagonizing multiple targets, such as dopamine receptors (e.g., D2 and D4) and serotonin receptors (e.g., 5-HT2A), clozapine effectively manages the multidimensional symptoms of schizophrenia while significantly attenuating the extrapyramidal side effects commonly associated with conventional single-target drugs. In oncology, sorafenib [22] represents a prototypical multi-kinase inhibitor that concurrently targets the RAS/RAF tumor cell proliferation signaling pathway and vascular endothelial growth factor receptor (VEGFR)–mediated tumor angiogenesis [23]. This dual mechanism exerts both antiproliferative and anti-angiogenic activities, thereby blocking activation of alternative signaling pathways and mitigating risks of drug resistance and disease progression. Similarly, in type 2 diabetes, dual PPARα/γ receptor agonists [24] simultaneously enhance insulin sensitivity and improve lipid metabolism, lowering blood glucose and triglyceride levels while avoiding the weight gain adverse effects often induced by selective PPARγ agonists [25]. These success stories underscore the advantages of multi-target drugs in enhancing therapeutic efficacy, overcoming drug resistance, simplifying treatment regimens, and addressing the complex mechanisms of multifaceted diseases. Consequently, polypharmacology has firmly established itself as a highly promising strategy in modern drug discovery. Representative clinical successes of the "one drug, multiple targets" strategy and their polypharmacological mechanisms are summarized in Table 1.

Multi-target drug design is inherently complex because different targets may possess distinct binding sites, dynamic conformations, and low sequence similarity. Classical computational methods, such as ligand-based QSAR or single-target molecular docking, often struggle to simultaneously address these multidimensional requirements. GDL provides a powerful new paradigm by leveraging geometric representations, such as atomic 3D coordinates, surface shapes, and interaction graphs, to learn structural features and relationships that are critical for multi-target efficacy. Deep learning models that incorporate complete protein 3D geometry are overcoming the limitations of earlier ligand-centric approaches by clarifying how ligand shape and orientation influence binding across multiple targets. Moreover, GDL's capacity to derive insights from structural data is expanding the boundaries of the druggable genome, enabling the identification of previously intractable new binding sites and targets.

### 2.2 Overview of classic GDL models in drug design

The progression of GDL in drug design reflects a fundamental shift from topology-centric graph representations to fully 3D, symmetry-aware frameworks capable of encoding the precise spatial complementarity required for polypharmacology. In multi-target settings, a single ligand must simultaneously satisfy the geometric constraints of multiple, often structurally distinct binding pockets, which places much greater demands on explicit 3D modeling than in single-target scenarios. Early graph neural networks provided essential graph-theoretic foundations but exposed critical shortcomings in handling rigid-body transformations and directional geometry. Subsequent architectures systematically addressed these gaps, culminating in equivariant and generative models that directly operate on atomic coordinates and protein–ligand co-graphs. This evolutionary trajectory, which extends from topology-based neighborhood aggregation to SE(3)-equivariant diffusion, underpins the pocket representation, latent-space comparison, and end-to-end multi-target generation strategies examined in later sections. Representative architectures are illustrated in Figure 2, with systematic comparisons in Table 2.

Early graph-based models, including Graph Convolutional Networks (GCNs) [26], Graph Attention Networks (GATs) [27], and the unifying Message Passing Neural Networks (MPNNs) [28], treat molecules and proteins as abstract topological graphs, with atoms or residues as nodes and bonds or spatial proximity as edges. By aggregating neighborhood features or assigning learnable attention weights, these architectures excell at capturing local connectivity patterns for molecular property prediction and residue-level binding-site classification, as exemplified by GrASP [29]. Yet their reliance on hand-crafted geometric descriptors (bond lengths, angles) to approximate 3D information renders them rotationally sensitive and poorly suited to multi-target scenarios, where ligand poses must remain consistent across arbitrarily oriented protein pockets. This limitation motivates distance- and angle-aware extensions such as SchNet [30] and DimeNet [31]. SchNet introduces continuous-filter convolutions over radial basis functions of interatomic distances, achieving global rotation and translation invariance without predefined bond graphs. DimeNet further incorporates three-body angular features, enabling directional message passing that preserves invariance under rigid transformations while capturing conformational nuances essential for quantum-mechanical property benchmarks. These invariant models mark the transition from purely topological to continuous geometric representations, providing a scalable foundation for atomic-level affinity prediction and early pocket scanning.

The next leap addresses a deeper symmetry requirement, demanding not merely invariance of scalar outputs but equivariance of intermediate vector and coordinate features under Euclidean transformations. E(n)-Equivariant Graph Neural Network (EGNN) [32] pioneers this by jointly updating node features and 3D coordinates via symmetry-preserving operations, such as distance-based coordinate updates, ensuring that rotated inputs produce correspondingly rotated outputs while scalar predictions remain invariant. Building on tensor-product mathematics, the SE(3)-Transformer [33] and Tensor Field Network (TFN) [34] rigorously encode higher-order spherical harmonics, allowing features to transform according to Wigner D-matrices. The Geometric Vector Perceptron Graph Neural Network (GVP-GNN) [35] offers a computationally lighter alternative by explicitly separating scalar (invariant) and vector (equivariant) channels, proving exceptionally effective for protein backbone modeling and functional-site prediction [36], as demonstrated by PocketMiner [13]. Collectively, these equivariant architectures reduce the reliance on data augmentation or auxiliary features and deliver superior generalization in protein–ligand binding tasks where absolute orientation is arbitrary, a property that is important for aligning multiple target pockets in polypharmacological design.

Task-oriented extensions translate these symmetry principles into practical SBDD pipelines. EquiBind [37] demonstrates end-to-end docking by encoding ligand and protein pocket graphs, fusing features via attention, and outputting an equivariant rigid-body transformation matrix, operating orders of magnitude faster than sampling-based methods while achieving high success rates on PDBbind. FRAME [38] advances iterative, structure-guided fragment growth using dual SE(3)-equivariant networks to predict attachment sites and optimal conformations directly from local protein–ligand geometry. Finally, GeoDiff [39] establishes geometric diffusion as a generative paradigm, defining forward noise processes in conformation space and training reverse models that respect internal-coordinate invariances, thereby enabling physically plausible sampling and subsequent pocket-conditioned extensions [40–42]. This progression, moving from topology to invariance, equivariance, and generation, equips GDL to address the combinatorial explosion and geometric constraints inherent in designing ligands that engage multiple targets simultaneously.

### 2.2.1 Invariance and equivariance in classic GDL models

Invariance denotes that a model’s output is insensitive to the choice of input coordinate frame. For example, models employing scalar features such as interatomic distances, bond angles, and dihedral angles (e.g., standard MPNNs, SchNet, and DimeNet) inherently exhibit rotational and translational invariance, meaning that the predicted molecular properties, such as energy, binding affinity, remain unchanged under any rigid-body transformation. These models achieve coordinate independence by incorporating relative geometric descriptors as inputs.

In contrast, equivariance imposes a stricter constraint, requiring intermediate representations and outputs to transform equivalently under rotations and translations of the input. For instance, EGNN updates node 3D coordinates such that a global rotation of the input induces an identical rotation in the output. Similarly, the SE(3)-Transformer and TFN rigorously ensure that internal features transform according to the Wigner D-matrices. This equivariant property is particularly valuable for tasks that predict geometric structures directly, such as molecular conformation generation or docking pose prediction, as the model must generate not only invariant scalars but also equivariant coordinates or vectors. Models such as EquiBind, DiffDock, and FRAME adopt equivariant architectures, ensuring that predicted ligand conformations or fragment positions co-transform with the protein–ligand coordinate system.

Notably, invariance represents a special case of equivariance, occurring when outputs are scalars. The choice between invariance and equivariance is task-dependent. Specifically, ensuring output invariance suffices for property prediction, whereas equivariance is essential for tasks requiring correctly oriented structural outputs, such as conformation generation and pose prediction. Generally, scalar-feature-based models such as MPNN, SchNet, and DimeNet are invariant, whereas EGNN, the SE(3)-Transformer, and GVP explicitly encode directional information, thereby making them equivariant. While some methods achieve approximate invariance via data augmentation (e.g., GCNs and GATs), directly embedding symmetry into the architecture yields greater robustness.

Crucially, in multi-target design, the reliance on purely invariant representations emerges as a critical bottleneck. The generation of polypharmacological ligands necessitates the simultaneous orientation of a molecule within two independent protein coordinate systems. Invariant models, relying solely on scalar distances, suffer from “directional blindnes” during 3D generation, as they cannot distinguish whether a growing moiety correctly fits target A or collides with the steric wall of target B. Equivariant architectures elegantly resolve this spatial dilemma by preserving absolute directional vectors. This enables the shared ligand to simultaneously process “directional geometric pulls” from multiple superimposed pockets by satisfying target A’s hydrogen-bonding trajectories while strictly avoiding target B’s steric boundaries, thereby providing the indispensable physical foundation for resolving complex multi-coordinate geometric conflicts.

### 2.2.2 Geometric modeling granularity: atomic-, residue-, and surface-level representations

GDL models can also be classified according to the granularity of their input data as atomic-, residue-, or surface-level models.

Atomic-level models construct graphs or point clouds directly with atoms as nodes, thereby preserving the fine-grained structural details of molecules. Most models for molecular property prediction and ligand generation (e.g., the GCN/GAT/MPNN family, SchNet, DimeNet and EGNN) belong to this category when all-atom features are provided. They operate on atomic graphs, such as the covalent-bond networks of small molecules or all-atom graphs of proteins. Their primary advantage lies in the completeness of information, which enables precise characterization of interactions. However, they incur higher computational costs for macromolecules such as proteins and struggle to directly capture macroscopic structural concepts. For instance, predicting protein pockets from atomic graphs typically requires post-processing clustering of atoms to identify pocket locations.

Residue- or backbone-level models represent amino acid residues or nucleotides as nodes. This coarser granularity reduces computational complexity and is particularly well suited to large-scale systems such as entire proteins [43]. A typical approach uses residue centroids or representative points as nodes, with edges defined by spatial proximity or sequential adjacency. Methods such as GraphBind and GraphSite employ residue-level graphs to predict binding sites or interfaces by framing the task as node classification (i.e., determining whether a residue belongs to a pocket). Although residue-level models omit intra-residue details and may thus achieve slightly lower precision, they offer practical advantages for global processing tasks (e.g., whole-protein pocket scanning). In protein–ligand affinity prediction, a common strategy employs a residue-level graph for the protein and an atomic-level graph for the ligand, with fusion achieved via attention or symmetry modules. This approach preserves computational efficiency while maintaining atomic resolution for the ligand component.

Surface-level models operate directly on protein surface point clouds or meshes, with an emphasis on surface geometry and physicochemical properties [44]. A representative example is MaSIF, which samples uniformly distributed points on the protein surface as nodes and precomputes features such as curvature, hydrophobicity, and charge. Local shape patterns are then extracted through convolutions on the surface mesh or point cloud, specifically along geodesic neighborhoods, to enable ligand or interface recognition. Surface-level models inherently exhibit rotational invariance, as surface descriptors are coordinate-independent, and they focus on the ligand-contacting interface, delivering superior performance in binding-site detection. This model efficiently identifies small-molecule binding patches and protein–protein interfaces. Its successor, dMaSIF, eliminates the need for precomputation, thereby enabling end-to-end learning of surface features. Other approaches (e.g., EquiPocket, PocketAnchor) fuse surface point clouds with internal atomic graphs, integrating geometric shape and amino acid information for pocket localization.

In summary, models of different granularities are suited to distinct scenarios. Specifically, atomic-level models excel at fine-grained small-molecule modeling and detailed protein–ligand interaction analysis, residue-level models are ideal for global feature extraction across entire proteins and rapid pocket screening, and surface-level models are optimal for interface shape matching and fast pocket localization [45]. In practice, these granularities can be combined, for example, by integrating protein surface point clouds with atomic graphs to leverage the strengths of each.

### 2.2.3 Algorithmic functional paradigms: discriminative and generative architectures

Finally, from a functional standpoint, GDL models can be broadly categorized into discriminative and generative paradigms.

Discriminative models aim to predict specific attributes or classes from the input data. In drug design, typical discriminative tasks include predicting the binding affinity of a given protein–ligand complex (regression), identifying viable binding pockets (binary classification or segmentation), and evaluating the quality of docking poses, among others. Most of the previously discussed models (e.g., GCN, GAT, MPNN, SchNet, DimeNet, EGNN, and GVP-GNN) can serve as discriminative models. For example, a GNN can encode the protein pocket and ligand into vector representations and subsequently predict their binding affinity. Alternatively, a GCN can assign a score to each residue, indicating its probability of belonging to a binding pocket. Discriminative models typically yield deterministic outputs without sampling from a data distribution. Their performance is evaluated using metrics such as accuracy and correlation coefficients.

In contrast, generative models aim to sample novel structures from the learned data distribution to fulfill specified objectives. In drug design, generative GDL models have garnered substantial attention in recent years for their potential to directly generate novel molecules or structures. These methods learn the underlying distribution of the training data and generate samples through processes such as decoding in variational autoencoders (VAEs) or reverse diffusion in diffusion models [46].

### 2.3 GDL methods for multi-target binding pocket representation and modeling

The critical first step in multi-target drug design is to understand and model the binding sites of distinct targets. GDL has catalyzed the development of innovative approaches that represent protein binding pockets in a more computationally efficient and interpretable manner, thereby enabling precise quantification of their similarities and differences. Early binding-site comparison methods typically relied on geometric templates, pocket alignment, or physicochemical descriptors, which are computationally demanding and often limited in interpretability. In contrast, GDL-based models, such as 3D convolutional neural networks (CNNs) and GNNs, directly learn pocket features from structural data. This facilitates both the identification of druggable pockets and the comparison of pocket similarities across proteins. For instance, Aggarwal et al. [47] introduced DeepPocket, a CNN-based framework that rescores and segments protein cavities pre-identified by geometric algorithms (e.g., Fpocket). By integrating classical pocket detection with deep learning, DeepPocket markedly improves the accuracy of binding-site ranking and boundary delineation, outperforming prior state-of-the-art methods on benchmark datasets. Importantly, rather than relying on hand-crafted features, these deep models learn which pocket shapes and chemical environments are more likely to be druggable, exhibiting robust generalization to novel structures.

Beyond identifying individual pockets, GDL methods further enable the association of binding sites across multiple targets [48]. Specifically, when two proteins share similar pocket geometries or interaction patterns, they are likely to bind the same ligand, which is a desirable outcome in polypharmacology design and a valuable signal for predicting off-target effects in safety assessments. Lohmann et al. [49] demonstrated that a GNN model for protein–ligand affinity prediction naturally learns a latent space in which binding-site embeddings cluster according to protein families and functions. This clustering indicates that the model's pocket representations capture essential geometric and functional features, thereby grouping targets amenable to modulation by similar ligands. The approach thus opens a "pocket-centric" route to multi-target modeling. For example, one can computationally assess whether the pocket embedding of a novel target lies near a known cluster, such as kinase ATP-binding sites, thereby inferring cross-target ligand applicability or supporting drug repositioning. The authors additionally observed that certain latent pocket clusters correspond to known functional families and that ligand size can shape their geometric distribution. These insights are highly valuable for multi-target design because they show that GDL models can determine whether two targets possess substantially similar binding sites, which is a prerequisite for a single agent to act on multiple targets simultaneously.

Concurrently, multi-task learning has been harnessed to predict binding sites for diverse ligand types. Yuan et al. [50] developed GPSite, which integrates Transformer-based protein language models with geometric features to predict binding residues across 10 ligand categories including DNA, RNA, peptides, proteins, small molecules (e.g., ATP or heme), and metal ions. The multi-task architecture is trained across multiple interaction types, forcing the model to extract protein geometric representations that generalize across modalities. Results show that GPSite substantially outperforms specialized sequence- or structure-based predictors on all tasks and remains robust even with AI-predicted protein structures. This underscores the capacity of geometric deep models to unify pocket modeling for diverse targets and ligands, which is an essential capability in polypharmacology, where targets range from enzymes and receptors to nucleic acid–binding proteins [51]. Similarly, Kozlovskii and Popov [52] recently noted that modern deep learning methods now represent a major category of binding-site identification tools, complementing traditional sequence-, template-, and physics-based approaches. These models routinely fuse sequence and structural information and handle small-molecule pockets, peptide interfaces, and metal-binding sites within a single unified framework.

In practice, GDL-driven pocket models enhance multi-target drug design in several key respects [53]. First, they accelerate the discovery of potential allosteric or secondary sites that enable simultaneous regulation of multiple pathways. This high-throughput, high-accuracy capability is especially vital for polypharmacology applications requiring genome-scale exploration. Second, vectorized latent pocket representations support rapid similarity searches. For example, if the pocket embedding of a novel pathogen protein closely matches that of a human enzyme target, it may flag cross-reactivity risks or repositioning opportunities. Third, these models readily incorporate pocket flexibility and contextual information. Specifically, frameworks such as 3D CNN-based segmentation or graph attention on interfaces not only predict binding locations but also highlight key residues, assisting medicinal chemists in elucidating structure–activity relationships within multi-target contexts. In summary, GDL provides faster and often more accurate tools for discovering and comparing binding sites under a pocket-centric paradigm than do traditional methods. These advances establish a robust foundation for designing ligands that optimally engage multiple pockets while avoiding undesired ones.

### 2.4 GDL-driven multi-target ligand generation and optimization

Designing a single molecule capable of simultaneously modulating multiple targets constitutes an intricate balancing act. The ligand must accommodate multiple binding sites that may differ markedly in spatial geometry and chemical environment while achieving ideal polypharmacological properties, including synergistic enhancement of target potency, optimization of physicochemical profiles, and minimization of off-target toxicity. Traditional *de novo* design methods, whether relying on rule-based fragment assembly or intuition-guided medicinal chemistry optimization, are highly prone to combinatorial explosion and spatial conflicts when simultaneously addressing the stringent 3D geometric constraints of multiple binding pockets.

In recent years, generative GDL has propelled multi-target drug design from blind exploration in ligand chemical space toward rational generation guided by multi-receptor geometric constraints, explicitly incorporating target 3D structural information [54–56]. This development has progressed from early fragment assembly based on local structural constraints to current end-to-end joint generation facilitated by equivariant diffusion models integrated with reinforcement learning (RL).

#### 2.4.1 Early explorations based on fragment assembly and structural conditioning

During the initial stages of multi-target design, GDL models primarily sought to inject geometric features from individual or partial protein pockets into the generation process and to explore 3D assembly rules at the fragment level. These efforts established a critical "pocket-aware" foundation for subsequent complex end-to-end multi-target generation.

Mapping to a 3D conditional feature space represents a pivotal step toward structure-guided generation. The RELATION model proposed by Wang et al. [57] constituted an early landmark. Employing an encoder–decoder architecture for 3D conditional generation, it learns a joint latent space between protein binding pockets and ligand shapes. This enables sampling of novel molecules exhibiting high complementarity to a specific target, guided by extracted pocket geometric features or pharmacophore constraints. Although RELATION primarily addresses single-pocket conditioning, the approach demonstrated the feasibility of using GDL to embed 3D context into molecular generation. Extending this concept to joint conditioning on multiple pocket features would allow generative models to directly identify candidate molecules at the intersection of multiple target feature sets.

To ensure 3D compatibility and drug-likeness of multi-target ligands, fragment-based deep geometric generation frameworks have gained increasing traction. Complex multi-target ligands can be conceptualized as assemblies of privileged pharmacophore modules serving distinct target binding sites. The FragGen system proposed by Zhang et al. [58] adopted a hybrid strategy combining deep learning with advanced geometric processing to assemble drug-like fragments in 3D space. Whereas early 3D generative models often yielded geometrically unrealistic conformations when fragments linking, FragGen decomposes the molecule's 3D construction into multiple geometric variables (e.g., torsion angles and bond lengths) and imposes physical constraints, thereby achieving reliable fragment-level generation. This mechanism ensures conformational rationality when appending modules in multi-target contexts.

Furthermore, for multi-objective optimization of lead compounds, Powers et al. [38] introduced the FRAME framework based on iterative growth. FRAME leverages SE(3)-equivariant GNNs to iteratively predict, starting from a given initial ligand–protein complex, where to attach a new fragment and which fragment conformation to add. The model internalizes 3D ligand expansion patterns and can simultaneously optimize affinity and selectivity while reducing off-target effects. In essence, this tightly coupled mechanism between local ligand geometry and protein constraints represents an embryonic form of automated balancing across multiple target optimization objectives in the high-dimensional polypharmacology design space.

Notably, diffusion generative models, which have revolutionized image generation, are now emerging in protein design and show promise for ligand design applications [59]. Watson et al. [60] demonstrated RFdiffusion, a diffusion model fine-tuned from a protein structure prediction network that can generate protein backbones and accomplish diverse design tasks ranging from monomers to enzyme active sites. The principal advantages of diffusion models, namely high output diversity and flexible conditioning during generation, are particularly attractive for drug design. In principle, such models can be conditioned simultaneously on multiple pocket grids or distance maps, allowing them to fuse geometric constraints from several targets throughout the progressive denoising process. Although small-molecule diffusion models remain in their infancy, promising preliminary results have appeared. By analogy with RFdiffusion's success in generating functional proteins experimentally validated via cryo-electron microscopy, researchers foresee that equivariant diffusion models for molecules hold strong potential to generate 3D conformations satisfying multi-target pharmacophore or shape constraints. Overall, GDL-driven molecular generation is transitioning from creating new molecules from scratch to structure-guided generation tightly coupled with target geometry. For multi-target design, this indicates that AI systems are approaching the ability to embed considerations for multiple binding sites and propose candidate drugs, substantially narrowing medicinal chemists' search space. Structure-guided generative models are now yielding molecules with higher predicted success rates in efficacy, novelty, and often superior binding metrics [61], providing a powerful foundation for discovering polypharmacological candidates.

### 2.4.2 GDL-driven generation of multi-target molecules

Although fragment growth and conditional generation hold considerable theoretical promise, traditional multi-target generation methods, such as RationaleRL [62] and MARS [63]), primarily rely on 1D sequences or 2D molecular fingerprints and lack the capacity for direct perception and resolution of 3D spatial conflicts [64]. In recent years, the maturation of equivariant diffusion models and their integration with RL have led to the emergence of algorithms capable of end-to-end and explicit handling of complex geometric conflicts in dual- or multi-target scenarios within vast chemical spaces. Table 3 provides a systematic comparison of the current state-of-the-art structure-aware multi-target ligand generation models, including AIxFuse [65], MDRL [66], and DualDiff [67]. As illustrated in Figure 3, these advanced GDL dual-target generation strategies can be distinctly classified into two paradigms based on their mechanisms for integrating multi-target information, namely “external reward induction” and “internal architecture coupling”.

The first paradigm employs external induction strategies based on physical simulation and RL, as exemplified by AIxFuse and MDRL. Given the relative scarcity of structural data for dual-target co-crystal complexes, the incorporation of external physical priors or scoring feedback has become instrumental in overcoming current limitations. As the first structurally aware dual-target generation model, AIxFuse innovatively integrates physical simulations such as molecular docking directly into the generation loop. The method employs two self-play Monte Carlo Tree Search (MCTS) agents to explore pharmacophore fusion patterns satisfying dual-target constraints, while utilizing the AttentiveFP graph attention mechanism to extract features for real-time estimation of dual-target affinity [68]. To overcome generalization bottlenecks, AIxFuse adopts an active learning strategy by iteratively incorporating molecular docking feedback to fine-tune the evaluation network. Experimental results demonstrate that this mechanism significantly enhances the structural adaptability of generated molecules, achieving multi-objective generation success rates on low-resource target pairs, such as RORγt/DHODH, substantially surpassing those of traditional data-driven models.

Building upon external constraint induction, Yuan et al. [66] proposed a phased MDRL framework combining a 3D diffusion model and RL. In the initial phase, MDRL employs a 3D diffusion model to learn a universal distribution of drug-like molecular point clouds in a target-agnostic setting. This stage innovatively replaces conventional MLPs with Kolmogorov–Arnold Networks (KANs), markedly improving the model's expressive capacity and interpretability, yielding chemical validity exceeding 99% for generated molecules. In the subsequent RL phase, MDRL integrates predicted docking affinities for two specific targets with drug-likeness properties, such as quantitative estimate of drug-likeness (QED), log $P$, and synthetic accessibility (SA), into a joint reward function. Through policy gradient optimization, the generation trajectory is guided toward Pareto-optimal solutions featuring both high dual-target affinity and superior physicochemical properties within vast chemical space. This "unsupervised generation and guided optimization" paradigm exhibits exceptional flexibility and effectively circumvents drug-likeness issues, such as excessive molecular volume arising from the sole pursuit of docking scores.

In contrast to methods reliant on external scoring and RL screening, the second paradigm achieves internal geometric coupling via equivariant message passing. As a prime example, Zhou et al. [67] proposed resolving dual-target geometric conflicts through compositional generation in diffusion models by directly embedding the solution into the model architecture. The core strategy entails rigid-body superposition of the binding pockets of the two proteins in 3D space, guided by the binding conformation of known anchor ligands, to construct a shared ligand–dual-target composite graph. On this composite graph, an SE(3)-equivariant message-passing network enables the ligand to simultaneously receive geometric constraint signals from both "locks" (target A and target B) throughout the reverse diffusion process. To determine the optimal fusion mechanism, the authors compared CompDiff, which merges gradients only at the output layer of the diffusion model, with DualDiff, which achieves deep fusion of messages from both targets at every layer. Results demonstrate that DualDiff, through its layer-wise information fusion, markedly outperforms its counterpart and more effectively captures the structural mechanisms underlying simultaneous ligand binding to both proteins. Moreover, this strategy establishes a highly valuable zero-shot transfer paradigm, enabling direct multi-target combination sampling with pre-trained single-target diffusion models without task-specific fine-tuning. Benchmark evaluations demonstrate that molecules generated by DualDiff comprehensively outperform those from traditional single-target generation followed by superposition or rigid fragment-linking algorithms in dual-target affinity and dual-high-affinity ratios. Compared with conventional "screen A then screen B" multi-step serial filtering or mechanically constrained fragment splicing based on prior knowledge, the latest end-to-end multi-target generation models powered by GDL exhibit a dimensionality-reduction advantage. The DualDiff model exemplifies the internal geometric coupling paradigm by directly resolving spatial conflicts between the two pockets at the foundational 3D level through shared coordinate systems and equivariant networks, thereby demonstrating exceptional geometric intuition. In contrast, MDRL and AIxFuse represent the external reward induction paradigm, leveraging RL and physical simulation feedback to implicitly accommodate multi-target demands while endowing the model with high flexibility in balancing multiple drug-likeness metrics. The emergence of these end-to-end generation frameworks signifies that polypharmacology ligand design has formally transitioned from serendipitous blind screening or multi-step splicing, thereby surmounting the computational barrier of combinatorial explosion and entering an era of rational automated generation driven by precise 3D structural constraints [69].

Building upon these foundational end-to-end frameworks, recent advancements aim to eliminate the residual geometric bottlenecks associated with pocket superposition. Although internal coupling methods such as DualDiff effectively resolve spatial conflicts in 3D space, their reliance on rigid pocket alignment introduces artificial constraints that may deviate from true polypharmacological binding distributions. To overcome this limitation, Wu et al. [70] introduced FuseDiff, a symmetry-preserving joint diffusion model that completely bypasses the need for global pocket alignment. Instead of superimposing the two binding pockets into a shared coordinate system, FuseDiff directly models the joint density of a shared 2D molecular graph and two independent, target-specific 3D binding poses. This is accomplished via a Dual-target Local Context Fusion mechanism, which enables each ligand atom to dynamically aggregate geometric contexts from both unaligned pockets during message passing. Furthermore, by simultaneously generating explicit chemical bonds, this model enforces stringent topological consistency between the two distinct conformations while accommodating target-specific spatial adaptations. The advent of such alignment-free, joint-density generative architectures represents a significant advancement in capturing the true physical complexity of multi-target molecular recognition.

### 2.5 Multimodal and structural data integration in polypharmacology

Multi-target drug design does not occur in an isolated structural vacuum, but rather lies at the intersection of multiple biological data modalities. Targets may be interconnected within signaling networks, drugs induce phenotypic responses, and genomic as well as proteomic backgrounds often determine why simultaneously modulating targets A and B proves more effective. GDL models are increasingly integrating diverse data modalities, including sequences, structures, networks, and even cellular-level data, to develop more comprehensive drug–target interaction predictors. In the context of polypharmacology, such integration is critical, as the required understanding extends beyond mere protein–ligand binding to include protein–protein interactions, pathway effects, and overall organismal outcomes.

A prominent example stems from the integration of knowledge graphs and ontologies [71,72]. Jeong et al. [73] proposed GeOKG, a geometrically aware knowledge graph embedding model that incorporates the hierarchical structure of the Gene Ontology (GO) along with gene annotations to enhance biological predictive capabilities. By fusing multiple geometric representations, including Euclidean and hyperbolic geometries, during training, GeOKG effectively captures the complex structures of GO terms and their interrelationships, transcending simple tree-like hierarchies. Its benefits are particularly pronounced in protein–protein interaction (PPI) prediction, where GeOKG outperforms earlier approaches. For multi-target design, this integration offers substantial value, as multi-target drugs frequently aim to perturb specific PPI subnetworks or pathway combinations. Protein representations generated by models such as GeOKG simultaneously encode functional annotations and network context, potentially allowing AI systems to discern that simultaneously targeting two proteins within the same GO pathway may elicit synergistic effects. This illustrates how graph deep learning merges ontological relationships with geometric reasoning to provide a more systemic perspective on biological targets.

Another key frontier in data integration concerns the fusion of chemical structural data with cellular and phenotypic information. The Hi-GeoMVP model [74], proposed by Chen and Zhang, exemplifies a quintessential multi-view approach. Hi-GeoMVP (Hierarchical Geometry-enhanced Multi-View Drug Response Prediction) integrates drug features, including molecular 2D graphs and 3D geometry, with multi-omics profiles of cancer cell lines to predict drug sensitivity ($IC_{50}$) across diverse cell lines. It seamlessly combines three modalities: (1) drug chemical structures encoded using geometry-enhanced GNNs, (2) cellular gene expression and mutation networks modeled via graph attention networks on gene interaction graphs, and (3) drug–cell line response data. Furthermore, Hi-GeoMVP employs a multi-task learning strategy, forecasting not only $IC_{50}$ values but also auxiliary outcomes such as cancer type and the drug's primary targeted pathways. This encourages the model to learn biologically meaningful features, such as associating drug structures with known target categories and linking cellular omics data with pathway activation. The results indicate state-of-the-art performance on the large-scale GDSC cell line panel [75,76], improving the Pearson correlation between predicted and observed sensitivity from 0.934 to 0.941, while demonstrating superior generalization in drug blind-testing scenarios. For multi-target drug development, models of this type offer transformative potential by facilitating *in silico* screening in diverse cellular contexts [77,78] while simultaneously accounting for molecular geometry and cellular genomics. For example, when a multi-target candidate is designed to modulate pathways such as EGFR and PI3K, Hi-GeoMVP-like models can predict its performance in tumors with varying pathway mutation backgrounds, thereby bridging target-level design with system-level outcome prediction.

Sequence–structure fusion represents another pivotal theme. Protein language models (pLMs), such as ESM and ProteinBERT, trained exclusively on sequence data, encode rich evolutionary information that complements geometric models [79–81]. A prominent example is SeaMoon, developed by Lombard et al. [82], which integrates a lightweight deep-learning module with a pLM to predict continuous protein motions (i.e., conformational deformations) solely from sequence information. SeaMoon's predictions of protein conformational heterogeneity complement purely geometry-driven normal-mode analysis by uncovering motion patterns overlooked by physical methods. This demonstrates that fusing sequence-derived embeddings, which capture evolutionary coupling information related to flexibility, with geometric modeling yields a more comprehensive characterization of target conformational landscapes. For multi-target drug design involving dynamic targets (e.g., allosteric sites or induced-fit binding), this fusion enables ligand designs based on realistic conformational ensembles rather than static structures. Similarly, Roche et al. [83] proposed ProRNA3D-Single, which merges embeddings from protein and RNA language models with geometric attention mechanisms to predict 3D structures of protein–RNA complexes from single-sequence inputs. Unlike methods relying on co-evolutionary information (multiple sequence alignments, MSAs), this approach leverages abundant sequence embeddings for pairwise alignment via geometric attention. In scenarios with sparse evolutionary data, it even surpasses AlphaFold 3 [83,84], underscoring the power of intelligent "sequence context and geometry" fusion in demanding interaction prediction tasks. In drug design, these methods facilitate predictions of interactions between proteins and non-coding RNAs or peptide hormones, thereby supporting modeling of multi-target mechanisms spanning diverse molecular entities. Cross-modal GDL is thus advancing predictions in multi-target scenarios that commonly involve multiple molecular classes.

Integration trends are also evident in multi-target functional prediction platforms. Yuan et al. [50] expanded GPSite into GPSFun, a web-based platform that unifies multiple GNN tasks, including ligand binding site prediction, metal-binding residue identification, GO function prediction, and subcellular localization, under the framework of "geometry-aware protein sequence functional annotation". By feeding pLM-derived sequence embeddings into geometric GNNs for structure-related tasks, GPSFun delivers comprehensive "one-stop" annotation across multiple functional categories. For drug designers, such platforms facilitate rapid profiling of novel targets by identifying functional sites and interaction partners, thereby establishing multi-target context (e.g., metabolite/DNA binding or interactions with known druggable partners) and helping determine whether a single-target strategy is adequate or a polypharmacology approach is warranted.

Finally, a distinctive dimension of data integration entails embedding AI predictions alongside physical or empirical rules within the design loop. A representative example is MEDICO proposed by Pang et al. [85], which incorporates docking scores as physically grounded priors during multi-view graph generation. Consequently, generated molecules appear plausible not only to the neural network but are also more physically compatible with target binding sites, effectively bridging deep learning with conventional structure-based design tools. This synergy produced substantial improvements in the discovery of SARS-CoV-2 main protease ($M^{pro}$) inhibitors, where optimized molecules exhibited approximately 88% higher docking affinity than their precursors. For multi-target design, multiple priors, such as docking to targets A and B alongside ADMET predictions, can be concurrently integrated into the generation loop, guided by composite objectives. In summary, multimodal integration ensures that multi-target drug design and target prediction screening no longer isolate individual targets. By interconnecting molecular structures [86,87], sequences [88], interaction networks [89,90], and phenotypic outcomes, these models enable systems-biology-driven assessment and design of multi-target interventions, which is a critical element in realizing efficacious and safe polypharmacology [91,92].

### 2.6 Validation and evaluation of multi-target models

With the advent of these advanced models, rigorous experimental validation and evaluation have become increasingly critical. Multi-target predictions inherently carry greater uncertainty, as candidate drugs must be assessed not only for binding to a single protein but also for their activity against multiple targets and their overall phenotypic effects. Researchers are employing diverse strategies to validate GDL-driven multi-target designs, ranging from computational benchmarks to *in vitro* experiments and, when necessary, *in vivo* studies. The following sections outline current approaches for evaluating these models and the compounds or predictions they produce.

### 2.6.1 Evaluation on established datasets and blind testing settings

Many GDL models are initially validated on widely recognized community benchmarks, typically employing cross-validation or held-out test sets. For example, binding site prediction models are commonly assessed on scPDB or PDB-derived datasets. DeepPocket [47] additionally incorporates the recent SC6K dataset of novel structures to assess generalization performance. Multi-task models such as GPSite and GPSFun have demonstrated superior performance compared to other methods in predicting binding residues for diverse ligand types across multiple datasets. In drug response prediction, Hi-GeoMVP has undergone extensive testing in drug-blind and cell-blind settings, predicting responses for previously unseen drugs or cell lines, and consistently outperforming previous models. These blind tests are particularly important for multi-target applications, as they better simulate real-world scenarios involving novel target combinations and contexts. A notable example is the evaluation by Qian et al. [93] in CAT-CPI model, which utilized not only standard protein–compound interaction datasets but also drug–drug interaction (DDI) datasets. The rationale was that accurate prediction of

compound–protein interactions should extend to identifying potential competition between two drugs for the same protein, a common DDI mechanism in polypharmacology. Their results highlighted the model's potential for DDI prediction, indirectly validating that its multimodal integration of molecular images and protein sequences effectively captures genuine interaction tendencies.

### 2.6.2 Prospective validation and experimental determination

The ultimate test for any drug design model lies in its ability to generate compounds that prove effective in the laboratory. Encouragingly, multiple GDL-designed molecules and proteins have been synthesized and experimentally characterized, providing concrete evidence of their translational potential. In the realm of protein design. For instance, RFdiffusion-generated influenza hemagglutinin-binding proteins were experimentally confirmed via cryo-electron microscopy, with structures closely matching the computational predictions. Furthermore, hundreds of designed proteins were tested for metal-binding and enzymatic activities, indicating that diffusion models can produce functional, and occasionally multifunctional, proteins. For small molecules, AI-designed compounds have attained nanomolar potency [94], as demonstrated in the FragGen study, where top-generated kinase inhibitors were experimentally validated, several exhibiting $IC_{50}$ values in the low nanomolar range alongside novel chemical scaffolds. These successes demonstrate that geometry-guided generative models can propose drug-like molecules amenable to experimental verification [95], extending beyond mere computational elegance. Likewise, MEDICO developed by Pang et al. [85] yielded novel SARS-CoV-2 $M^{pro}$ inhibitors. Although not all were synthesized, docking and molecular dynamics simulations indicated substantially enhanced binding affinities relative to known leads. These outcomes bolster confidence that incorporating docking scores within the generative framework produces molecules whose computational behaviors more closely align with genuine binders. In the near future, it is expected that GDL-designed multi-target compounds will undergo experimental validation, confirming their capacity to engage multiple targets simultaneously and elicit the intended polypharmacological effects. For instance, a single molecule inhibiting two kinases could achieve superior anti-tumor efficacy compared to individual inhibition.

### 2.6.3 Retrospective validation and case studies

Another validation strategy employs retrospective analyses to evaluate whether models can explain or rediscover known multi-target drugs and their interactions. For example, pocket similarity models can be assessed for their ability to cluster targets recognized by known broad-spectrum drugs. A latent space analysis by Lohmann et al. [49]

demonstrates that proteins modulated by similar ligands tend to cluster together. Similarly, multi-omics models such as Hi-GeoMVP can be probed by inputting known multi-target drugs to determine whether the model accurately predicts heightened sensitivity in cell lines with specific pathway activation backgrounds. The original study reveals that their multi-view design markedly enhanced prediction accuracy for drugs exhibiting multi-mechanistic profiles. Such retrospective evaluations help establish confidence that the models genuinely capture authentic polypharmacological patterns. Moreover, the models' attention weights or saliency maps can sometimes illuminate the rationale underlying predictions. For instance, they may highlight which specific molecular substructures drive selectivity for one target while preserving affinity for another. These interpretability features are particularly valuable to medicinal chemists in translating AI-generated insights into rational molecular design.

### 2.6.4 Systems-level validation

When multi-target models incorporate systems-level data, such as pathways, protein–protein interactions, and omics profiles, validation can extend to examining downstream biological effects. Integrated models may predict that simultaneous inhibition of targets A and B will induce apoptosis in cancer cells based on gene network analysis. These predictions can be experimentally confirmed by monitoring apoptosis markers during dual-target inhibition, whether through multi-target compounds or combination therapies. Although individual papers seldom provide comprehensive systems-level validation, research in systems pharmacology is progressively advancing this frontier. A notable example is the work by Guo et al. [96], who employed deep learning to design proteins featuring switch-like conformational changes, with the ligand-induced switching behavior subsequently confirmed through multiple crystal structures. Their analytical workflow integrated structural elucidation, molecular simulations, and mutational studies, providing a valuable template for evaluating whether drugs effectively modulate multiple conformational states or targets as intended. This highlights the critical reality that multi-target efficacy frequently depends on engaging the appropriate target combination in the correct conformational state. As such, advanced structural characterization methods, including the resolution of co-crystal or cryo-EM structures of a drug in complex with targets A and B individually, are becoming the gold standard. Indeed, several research groups have already utilized these high-resolution structural techniques to confirm drug binding to macromolecular complexes or large assemblies, thereby verifying that the compounds act on all intended components as designed.

### 2.6.5 Public resources and community evaluation

Community-shared models and datasets further facilitate broader validation efforts. Tools such as DeepPocket [47] and GPSFun [50] have been made available as open-source software or web services, enabling researchers to apply them to their specific multi-target challenges and report performance outcomes. This open accessibility accelerates validation across diverse contexts. For example, pharmaceutical companies may integrate RELATION [57] into their in-house multi-target drug discovery pipelines, with initial internal reports on AI-designed clinical candidates now emerging. Furthermore, community competitions, exemplified by the CASP protein–ligand prediction challenges and D3R drug design challenges [97], often implicitly involve multi-target considerations, such as ligand pose prediction across homologous proteins or multi-target affinity ranking. GDL models have demonstrated strong performance in these challenges, frequently achieving top rankings in single-target benchmarks and thereby establishing a solid foundation for the development of dedicated multi-target evaluation frameworks.

In summary, the validation of GDL-based approaches for multi-target drug design necessitates both rigorous computational benchmarking and real-world experimental confirmation. These models not only match or outperform prior methods on established test sets but also yield molecules and proteins that prove feasible and effective in experimental settings. For polypharmacology, this empirical validation is particularly crucial, providing compelling evidence that AI systems can successfully navigate the intricate design spaces traditionally dependent on the intuition and expertise of medicinal chemists. With ongoing testing and iterative refinement, continued improvements in predictive reliability are anticipated, thereby advancing AI-designed multi-target therapeutics toward preclinical development. Ultimately, the convergence of robust validation strategies, expanding high-quality datasets, and continuous model enhancement will determine the credibility of GDL in polypharmacology, guiding these models to prioritize the key molecular features that govern multi-target efficacy and patient safety [98].

### 2.7 Interpretability analysis of GDL-driven multi-target generative models

A primary challenge in multi-target ligand design is that a single ligand must accommodate the spatial structures of two distinct protein binding pockets, potentially requiring conformational adjustments in the ligand or the proteins themselves. GDL models offer a distinct advantage by directly modeling ligand–protein interactions in 3D space, thereby enabling explicit evaluation of how effectively a ligand satisfies the geometric constraints of both pockets. Zhou et al. [67] aligned the two target binding pockets in 3D space, incorporate prior knowledge of established protein–ligand interactions, and construct a dual-complex graph featuring a shared ligand node. They then integrate signals from both targets via SE(3)-equivariant message passing. This composite mechanism allows the ligand to receive geometric constraints simultaneously from both proteins, explicitly encoding the one-key-fits-two-locks binding paradigm.

This modeling approach confers substantial interpretability benefits. Researchers can trace messages received by individual ligand nodes from each protein graph to infer which ligand segments primarily interact with which target. The shared ligand graph decomposes binding contributions, clarifying which functional groups target protein A versus protein B. In contrast to simple late fusion of the two target representations, layer-wise message integration provides deeper insight into the simultaneous binding mechanism. Empirical results show that message combination at every layer (DualDiff) substantially outperforms output-layer fusion only (CompDiff), demonstrating that progressive integration of geometric information more effectively produces ligands compatible with both targets. From an interpretability perspective, the model incrementally incorporates constraints from both targets layer by layer, mirroring the iterative balancing process used by medicinal chemists in dual-target inhibitor design.

Moreover, SE(3)-equivariant diffusion models inherent to GDL

frameworks ensure invariance to spatial rotations and translations. Consequently, learned features depend solely on relative geometric relationships between ligands and proteins, independent of the absolute coordinate system. This symmetry property strengthens both physical plausibility and interpretability by ensuring that model predictions remain consistent regardless of spatial orientation, eliminating artifacts from arbitrary reference frames. Researchers can therefore confidently interpret the model as capturing authentic spatial complementarity, such as shape complementarity and electrostatic interactions, rather than spurious, frame-dependent features. Equivariant architectures, which operate on physically meaningful quantities such as interatomic distances, often internalize interpretable geometric rules, such as interaction thresholds based on distance. In this manner, GDL models extract principles akin to the classic “lock-and-key” paradigm of molecular recognition.

Notably, the incorporation of KANs [99] in generative models further advances interpretability. Conventional diffusion models rely on opaque MLPs for scoring functions. Yuan et al. [66] replaced these with KANs, achieving simultaneous gains in performance and transparency. Unlike standard MLPs, KANs place learnable activation functions along network edges as parameterized splines. This design, grounded in the Kolmogorov–Arnold representation theorem, decomposes complex multivariate functions into sums of simpler univariate functions, making each contribution highly interpretable. For instance, individual edge functions can illustrate how interatomic distance modulates generation probability, with peak contributions within specific ranges and declining beyond defined thresholds. Empirical evidence indicates that KAN-based models match or surpass MLP performance while offering substantially greater mechanistic insight. Researchers can thus elucidate how the model has internalized established physicochemical principles (e.g., effects of bond length or hydrophobicity) and applied them to dual-target design. Overall, KAN integration exemplifies how architectural refinements can markedly improve the transparency of generative processes in GDL.

Although prevailing GDL models assume rigid protein structures, they nevertheless provide valuable insights into conformational flexibility. By comparing the poses of generated ligands across the two binding environments, researchers can pinpoint ligand regions requiring torsional or rotational adjustments to accommodate pocket differences. Near-identical poses predicted for targets A and B suggest either highly similar pocket geometries or sufficient intrinsic ligand flexibility. Because GDL directly operates on 3D coordinates and distances, it readily accommodates incorporation of conformational ensembles. Training on multiple protein conformations or flexibility-induced perturbations enables models to learn robustness to minor pocket deformations, yielding ligands tolerant of structural variations. When a model generates consistent ligands across conformations, it indicates successful capture of the essential shared features between targets, which constitute the core determinants of effective dual-target engagement.

## 3. Data infrastructures for geometric polypharmacology

To advance GDL beyond conventional single-target SBDD, models must be trained on datasets capable of capturing the complexity of multi-target interactions. Unlike classic affinity datasets such as PDBbind [100,101], which address only "one ligand – one pocket" scenarios, polypharmacology requires datasets that provide combinatorial supervision signals for training models on multi-body geometric systems.

The primary datasets fall into two major categories. High-throughput screening matrices supply dense training tensors for deep geometric encoders, whereas clinical knowledge bases furnish sparse yet authentic validation graphs.

### 3.1 High-throughput screening matrices: the synergy landscape

These datasets provide continuous dose–response matrices, delivering the high-volume dense supervision signals essential for pre-training large-scale geometric encoders.

### 3.1.1 DrugComb: data portal and analysis platform

In constructing the infrastructure for multi-target geometric design, the DrugComb database [102] serves as a foundational resource, analogous to ImageNet in computer vision. This platform integrates and standardizes data from over 35 independent screening projects and provides unified synergy scores for nearly 700,000 drug combinations across metrics such as ZIP, Bliss, Loewe, and HSA. For GDL, these standardized dose–response tensors enable models to learn complete interaction response surfaces rather than merely fitting single scalar values [103–105]. By employing target deconvolution to map combination drugs to their respective 3D target conformations, GDL models function as geometric compatibility scorers. Driven by massive data volumes, they implicitly learn how the stereochemical features of two independent ligands antagonize or synergistically enhance each other at the systemic network level.

### 3.1.2 NCI-ALMANAC: anticancer drug combination matrix

Curated by the U.S. National Cancer Institute, NCI-ALMANAC [106] represents the largest and most experimentally consistent drug combination screening dataset to date. It encompasses more than 290,000 synergy measurements for approximately 100 approved anticancer drugs across the NCI-60 cell line panel. Owing to its exceptional consistency, it serves as the gold standard for structural benchmarking. Given that most of these drugs possess well-defined targets and available PDB structures, the dataset is ideally suited for target–target interaction (TTI) learning. Researchers can explicitly construct geometric graphs of pocket A–pocket B pairs and supervise the model using synergy scores to learn allosteric or functionally complementary relationships between protein surfaces.

### 3.1.3 AstraZeneca–Sanger DREAM Challenge dataset

Rigorous out-of-distribution (OOD) generalization benchmarks are essential to verify whether geometric models truly capture the underlying 3D physical principles of multi-target synergy rather than merely memorizing statistical patterns from known 2D chemical structures. Originally released for community competitions, the AstraZeneca–Sanger DREAM Challenge dataset [107] offers an ideal testing platform. It comprises approximately 11,500 highly curated drug combinations with comprehensive multi-omics and molecular phenotypic annotations. Its primary strength lies in predefined splitting schemes designed specifically for OOD evaluation. Testing model performance on novel molecular combinations featuring unseen chemical scaffolds allows robust assessment of the model's inductive reasoning capabilities based on spatial structural mechanisms, thereby distinguishing it from pseudo-structural models reliant solely on transductive memorization of 2D chemical identities.

### 3.2 Context-aware and clinical benchmarks

Before exploring computational paradigms for multi-target drug design in depth, it is crucial to distinguish clearly between drug combination synergy and single-molecule polypharmacology. The former depends on complementary interactions among multiple independent molecules at the systemic network level and involves complex DDIs and uncontrollable additive toxicity. The latter requires a single ligand to simultaneously satisfy the stringent 3D spatial constraints of multiple targets. Although these approaches differ fundamentally in physical implementation, the foremost challenge in rational single-molecule multi-target drug design lies in identifying which target pairs should be modulated by the same molecule. In this context, large-scale drug combination screening datasets, such as DrugComb, are being innovatively repurposed as system-level priors to guide GDL models toward protein pocket combinations with genuine synergistic therapeutic potential, thereby bridging macroscopic pathway synergy with microscopic geometric adaptation.

#### 3.2.1 SYNERGxDB: integrated pharmacogenomics database

Pharmacological synergistic effects are frequently context-dependent,

as the same target combination may produce markedly different phenotypes under varying genomic mutations or transcriptional backgrounds. Pure 3D structural and chemical screening matrices often overlook these microenvironmental differences. SYNERGxDB [108] fills this gap by aligning large-scale drug combination screening results with the genomic and transcriptomic profiles of specific cell lines. For multimodal GDL approaches, this database is indispensable for heterogeneous graph learning. Using SYNERGxDB, researchers can train environmentally aware conditional GNNs that explicitly condition predictions of geometric compatibility between receptors A and B on disease-state-specific dynamic gene expression networks [109]. This framework, integrating 3D structural data with high-dimensional omics [110,111], enables algorithms not only to assess spatial physical compatibility but also to ensure that dual-target interventions generate genuine synergistic effects within specific pathological networks, thereby supporting highly personalized multi-target pharmacological modeling.

### 3.2.2 DCDB 2.0 and CDCDB: clinical ground truth

High-throughput *in vitro* screening data primarily capture coarse-grained cell viability phenotypes that are often confounded by non-specific additive toxicity. This confounding effect creates a substantial translational gap between early *in vitro* metrics and real-world therapeutic efficacy in complex human systems. To guarantee that AI-recommended target combinations are truly safe and effective, real-world clinical data must serve as rigid constraints. The Drug Combination Database (DCDB 2.0) [112] systematically compiles verified successful combination regimens from clinical trials and the FDA Orange Book, while the Continuous Drug Combination Database (CDCDB) [113] provides ongoing incremental updates drawn from ClinicalTrials.gov and patent literature. In GDL-driven computational frameworks, these clinical databases function as a compass, supplying high-confidence positive edges for generative knowledge graphs. Supervision from these authentic clinical anchors enables rigorous validation of whether GDL-inferred target combinations possess substantive clinical feasibility and systemic safety, rather than merely demonstrating strong *in vitro* cytotoxicity. Moreover, CDCDB's continuous updates support dynamic benchmarks for temporal link prediction, allowing models to perform graph evolution training on historical clinical data streams and prospectively predict novel multi-target synergistic mechanisms likely to achieve future regulatory approval. Table 4 provides a systematic summary and comparison of the publicly available drug combination datasets currently applicable to geometric polypharmacology and multi-target drug design.

### 3.3 Methodology: construction of target–target geometric benchmarks

To systematically translate macroscopic drug combination synergy into microscopic 3D geometric tasks suitable for GDL, the field necessitates a standardized workflow. Figure 4 systematically outlines a comprehensive conceptual framework for constructing robust target-target benchmarks, encompassing the entire pipeline from phenotype deconvolution to multi-objective geometric evaluation.

#### 3.3.1 Data foundation: mapping synergistic phenotypes to 3D structures

Although conventional drug combination databases such as DrugComb provide extensive dose–response matrices, they lack the direct 3D protein structural information required for GDL. Recently, Zhou et al. [67] made foundational contributions to bridging this phenotype–structure gap. Starting from DrugCombDB, they screened drug combinations exhibiting positive synergistic effects using four mainstream synergy scoring systems, namely ZIP, Bliss, Loewe, and HSA. They subsequently performed target deconvolution to map the drugs to specific protein targets, and integrated PDBbind crystal structures with the AlphaFold database via P2Rank and AutoDock Vina to obtain high-confidence complex conformations. This process resulted in a dataset of 12,917 pairs of synergistically positive target combinations, establishing a standardized structural foundation for multi-target scenarios.

### 3.3.2 Task elevation: construction of target–target geometric learning benchmarks

Pharmacological interventions frequently exhibit polypharmacology and drug promiscuity, in which a single drug molecule is often mapped to multiple potential targets. When high synergistic phenotypes are observed for a drug pair, directly performing pairwise mapping of their respective candidate target sets and leads to severe combinatorial explosion and substantial false-positive noise from non-core "bystander targets".

To precisely identify the target pairs that genuinely drive the observed phenotypic synergy, we propose constructing a global "Weighted Target Interaction Graph" based on the foundational annotations provided by Zhou et al. [67]. Macroscopic drug synergy scores (e.g., ZIP, Bliss) are projected onto this network for graph topology-based score deconvolution and credit assignment. Under cross-validation from extensive multi-drug combination evidence, "driver target pairs" with genuine biological value for joint intervention accumulate high edge weights, effectively filtering incidental off-target noise.

To establish a predictive benchmark suitable for discriminative GDL, topological contrastive sampling is recommended, whereby highest-confidence core edges serve as positive samples, while distant node pairs lacking connectivity or showing explicit antagonism serve as negative samples . This approach bridges the semantic gap between macroscopic multi-target phenotypes and microscopic purified target pairs, providing a balanced benchmark that incorporates systems biology priors for 3D geometric discriminative tasks.

## 4. Challenges and future perspectives

The conventional "one drug, one target" paradigm in SBDD has revealed critical therapeutic limitations, particularly its propensity to elicit compensatory signaling pathways when addressing complex systemic diseases such as cancers exhibiting drug resistance, neurodegenerative disorders, and metabolic syndromes. In contrast, polypharmacology advocates a "one drug, multi-target" strategy that offers a promising paradigm shift for managing multifaceted diseases. Nevertheless, the rational design of a single ligand capable of simultaneously satisfying the stringent 3D spatial constraints of multiple targets within vast chemical space remains a formidable computational challenge, compounded by combinatorial explosion and intricate geometric conflicts.

This review systematically delineates how GDL overcomes these limitations, transitioning multi-target drug design from serendipitous blind screening or rudimentary fragment linking toward rational, structure-driven molecular generation. Progressing from the early vectorization of 3D binding pocket features and clustering of potential allosteric sites, through equivariant fragment growth, to state-of-the-art end-to-end dual-target generative architectures integrating SE(3)-equivariant diffusion models, RL, and KANs, GDL has demonstrated substantial potential in analyzing and resolving complex inter-receptor geometric conflicts. Consequently, drug discovery is entering a new era of multi-target molecular engineering guided by principles of spatial symmetry and fundamental physical laws.

Despite the revolutionary prospects of GDL in multi-target drug design, the field remains in its nascent stage. To successfully translate *in silico* prototypes into clinical candidates, several major challenges must be addressed:

(1) Transcending the static rigid-body assumption through the integration of macromolecular flexibility and dynamic conformational modeling. Most contemporary state-of-the-art dual-target generation models, including DualDiff, assume rigid protein structures and rely on rigid-body superposition to compute spatial compatibility. However, molecular recognition is inherently dynamic, and polypharmacology frequently depends on conformational flexibility and induced-fit mechanisms, whereby a single ligand can induce varying degrees of side-chain rearrangements or backbone shifts across different targets. Future GDL frameworks will need to advance from static “3D” to dynamic “4D” modeling [114] by augmenting 3D space with temporal dynamics or conformational ensembles. This can be achieved by incorporating molecular dynamics (MD) simulation trajectories or flexibility features derived from large language models into the reverse denoising process of diffusion models [115]. Such advances will enable the generation of multi-target ligands that are robust to local pocket deformations.

(2) Escaping the property trap to achieve Pareto optimality between multi-target affinity and drug-likeness properties. During generation, models attempting to satisfy multiple geometric constraints are prone to indiscriminately increasing molecular size and hydrophobicity to maximize binding energies. This frequently yields oversized molecules with excessively high lipophilicity (log *P*), resulting in elevated pharmacokinetic failure risks. Future generative models will require advanced multi-objective optimization frameworks. By integrating structurally interpretable architectures such as KANs with guided diffusion and RL, these models can navigate the complex Pareto front encompassing binding affinity, QED, SA, and off-target toxicity mitigation, thereby ensuring that the generated compounds are not only geometrically compatible but also genuinely drug-like.

(3) Decoupling phenotypic synergy from physical geometry to overcome affinity fixation. Although synergy matrices from drug combination screens provide valuable guidance, they introduce substantial

PK/PD confounding. Unlike combination therapies, which permit dynamic dose-ratio adjustments, single-molecule multi-target agents possess affinity ratios inherently fixed by chemical structure upon generation. Current models frequently maximize dual-target binding energies at the expense of biologically relevant asymmetric inhibition preferences and cannot equate *in vitro* combination ratios to single-molecule affinities due to baseline differences between compounds. A promising solution is occupancy-driven conditional generation, whereby quantitative systems pharmacology (QSP) models derive rational target occupancy (TO) distributions from phenotypic synergy data to serve as geometric constraints. These distributions then guide diffusion models in producing single-molecule candidates that precisely match the required occupancy ratios, thereby achieving accurate decoupling of phenotypic outcomes from physical geometry.

(4) Bridging the dry–wet gap through automated closed-loop validation from virtual generation to experimental testing. Computational predictions, such as Vina scores or MM/GBSA calculations, require rigorous experimental validation in the physical world. At present, GDL-generated multi-target candidates lack extensive wet-lab confirmation, and high virtual scores do not necessarily correspond to authentic intracellular target occupancy or multi-pathway therapeutic efficacy. The future of computational polypharmacology lies in establishing fully integrated dry–wet closed loops. These systems will leverage automated robotic synthesis platforms , known as self-driving laboratories [116], and high-throughput microfluidic phenotypic screening [117] to rapidly synthesize and evaluate AI-designed candidates. Real-time feedback of experimental results into GDL models via active learning will continuously refine scoring functions and generation strategies, serving as the critical bridge toward clinical translation.

(5) Integrating customized "negative design" to overcome translational bottlenecks. Off-target toxicity remains the most formidable barrier to the clinical translation of small-molecule therapeutics, thereby endowing anti-target avoidance strategies with substantial industrial value. Critically, the requisite anti-target panels are highly project-specific. For instance, a multi-target oncology compound must evade a distinctive set of structurally homologous proteins, whereas a central nervous system (CNS) candidate encounters a disparate array of receptor liabilities, in addition to ubiquitous safety threats such as the cardiotoxic hERG channel and CYP450 enzymes. Nevertheless, current GDL models predominantly maximize on-target affinities through "positive design", frequently yielding clinically unviable "dirty drugs". Future frameworks will need to explicitly incorporate customizable "negative spatial constraints". By embedding project-specific repulsive gradients into the diffusion process, algorithms can deliberately engineer sterically bulky moieties that achieve optimal complementarity with therapeutic binding pockets while inducing targeted steric clashes exclusively in the designated anti-targets. Mastery of this context-aware subtractive geometry is indispensable for converting computational hits into commercially viable clinical candidates.

From the pursuit of precise “magic bullets” targeting individual proteins to the engineering of “magic shotguns” capable of modulating complex biological networks, polypharmacology embodies a higher-dimensional strategy for confronting multifaceted disease mechanisms. Powered by its unparalleled 3D spatial awareness, GDL has illuminated the microscopic geometric principles of multi-target molecular design. By deeply integrating physical priors, cross-modal data [118], and fully closed dry–wet experimental loops, the next generation of AI-designed therapeutics with superior polypharmacological profiles is poised to bridge the “valley of death” in translational medicine [119] and deliver transformative solutions for major human diseases.

**CREDIT author statement**

**Tianming Han**: Data curation, Investigation, Methodology, Software, Writing – original draft. **Zhijie Pan**: Visualization, Validation. **Wenchi Ge**: Visualization, Validation. **Qi Zhao**: Conceptualization, Funding acquisition, Methodology, Project administration, Supervision, Writing – original draft, Writing – review & editing.

## Funding

This study is supported by Science and Technology Plan Project of Liaoning Province (Grant No. 2025-MSLH-351), Fundamental Research Funds for the Liaoning Universities (Grant No. LJ212410146026).

## Conflicts of interest

The authors declare that the research was conducted in the absence of any commercial or financial relationships that could be construed as a potential conflict of interest.

## 5. References

[1] Kabir A, Muth A. Polypharmacology: The science of multi-targeting molecules. Pharmacological Research. 2022;176:106055. doi:10.1016/j.phrs.2021.106055.
[2] Proschak E, Stark H, Merk D. Polypharmacology by Design: A Medicinal Chemist's Perspective on Multitargeting Compounds. Journal of Medicinal Chemistry. 2019;62(2):420–444. doi:10.1021/acs.jmedchem.8b00760.
[3] Munson BP, Chen M, Bogosian A, Kreisberg JF, Licon K, Abagyan R, et al. De novo generation of multi-target compounds using deep generative chemistry. Nature Communications. 2024;15(1):3636. doi:10.1038/s41467-024-47120-y.
[4] Sun L, Wang L, Guo D, Liu X, Wang B, Zhao Y, et al. Cucurbitacin B mitigates Staphylococcus aureus pathogenicity and reprograms macrophage responses to restore host defense. Journal of Pharmaceutical Analysis. 2026;16(5):101557. doi:10.1016/j.jpha.2026.101557.
[5] Zhou L, Ju Y, Cao Z, Cai S, Su J, Lu J. DNA-encoded library screening identifies CDK2-targeting lead compounds with favorable drug-like properties for anticancer development. Journal of Pharmaceutical Analysis. 2026;16(5):101498. doi:10.1016/j.jpha.2025.101498.
[6] Wei W, Li Z, Wang B, Liu Y, Sun Y, Wen D, et al. Pharmacological effects, classification, genetic and molecular studies of different chemotypes essential oil of Perilla frutescens (L.) Britt.: A review. Journal of Pharmaceutical Analysis. 2026;16(5):101454. doi:10.1016/j.jpha.2025.101454.
[7] López-López E, Medina-Franco JL. Toward structure–multiple activity relationships (SMARts) using computational approaches: A polypharmacological perspective. Drug Discovery Today. 2024;29(7):104046. doi:10.1016/j.drudis.2024.104046.
[8] Cichońska A, Ravikumar B, Rahman R. AI for targeted polypharmacology: The next frontier in drug discovery. Current Opinion in Structural Biology. 2024;84:102771. doi:10.1016/j.sbi.2023.102771.
[9] Isert C, Atz K, Schneider G. Structure-based drug design with geometric deep learning.

Current Opinion in Structural Biology. 2023;79:102548. doi:10.1016/j.sbi.2023.102548.
[10] Hekkelman ML, De Vries I, Joosten RP, Perrakis A. AlphaFill: enriching AlphaFold models with ligands and cofactors. Nature Methods. 2023;20(2):205–213. doi:10.1038/s41592-022-01685-y.
[11] Abramson J, Adler J, Dunger J, Evans R, Green T, Pritzel A, et al. Accurate structure prediction of biomolecular interactions with AlphaFold 3. Nature. 2024;630(8016):493–500. doi:10.1038/s41586-024-07487-w.
[12] Varadi M, Anyango S, Deshpande M, Nair S, Natassia C, Yordanova G, et al. AlphaFold Protein Structure Database: massively expanding the structural coverage of protein-sequence space with high-accuracy models. Nucleic Acids Research. 2022;50(D1):D439–D444. doi:10.1093/nar/gkab1061.
[13] Meller A, Ward M, Borowsky J, Kshirsagar M, Lotthammer JM, Oviedo F, et al. Predicting locations of cryptic pockets from single protein structures using the PocketMiner graph neural network. Nature Communications. 2023;14(1):1177. doi:10.1038/s41467-023-36699-3.
[14] Nippa DF, Atz K, Hohler R, Müller AT, Marx A, Bartelmus C, et al. Enabling late-stage drug diversification by high-throughput experimentation with geometric deep learning. Nature Chemistry. 2024;16(2):239–248. doi:10.1038/s41557-023-01360-5.
[15] Comajuncosa-Creus A, Jorba G, Barril X, Aloy P. Comprehensive detection and characterization of human druggable pockets through binding site descriptors. Nature Communications. 2024;15(1):7917. doi:10.1038/s41467-024-52146-3.
[16] Isigkeit L, Hörmann T, Schallmayer E, Scholz K, Lillich FF, Ehrler JHM, et al. Automated design of multi-target ligands by generative deep learning. Nature Communications. 2024;15(1):7946. doi:10.1038/s41467-024-52060-8.
[17] Méndez-Lucio O, Ahmad M, Del Rio-Chanona EA, Wegner JK. A geometric deep learning approach to predict binding conformations of bioactive molecules. Nature Machine Intelligence. 2021;3(12):1033–1039. doi:10.1038/s42256-021-00409-9.
[18] Du Y, Jamasb AR, Guo J, Fu T, Harris C, Wang Y, et al. Machine learning-aided generative molecular design. Nature Machine Intelligence. 2024;6(6):589–604. doi:10.1038/s42256-024-00843-5.
[19] Guo J, Fialková V, Arango JD, Margreitter C, Janet JP, Papadopoulos K, et al. Improving de novo molecular design with curriculum learning. Nature Machine Intelligence. 2022;4(6):555–563. doi:10.1038/s42256-022-00494-4.
[20] Chen Z, Peng B, Zhai T, Adu-Ampratwum D, Ning X. Generating 3D small binding molecules using shape-conditioned diffusion models with guidance. Nature Machine Intelligence. 2025;7(5):758–770. doi:10.1038/s42256-025-01030-w.
[21] Roth BL, Sheffler DJ, Kroeze WK. Magic shotguns versus magic bullets: selectively non-selective drugs for mood disorders and schizophrenia. Nature Reviews Drug Discovery. 2004;3(4):353–359. doi:10.1038/nrd1346.
[22] Wilhelm SM, Adnane L, Newell P, Villanueva A, Llovet JM, Lynch M. Preclinical overview of sorafenib, a multikinase inhibitor that targets both Raf and VEGF and PDGF receptor tyrosine kinase signaling. Molecular Cancer Therapeutics. 2008;7(10):3129–3140. doi:10.1158/1535-7163.MCT-08-0013.
[23] Cohen P, Cross D, Jänne PA. Kinase drug discovery 20 years after imatinib: progress and future directions. Nature Reviews Drug Discovery. 2021;20(7):551–569. doi:10.1038/s41573-

021-00195-4.
[24] Kaul U, Parmar D, Manjunath K, Shah M, Parmar K, Patil KP, et al. New dual peroxisome proliferator activated receptor agonist—Saroglitazar in diabetic dyslipidemia and non-alcoholic fatty liver disease: integrated analysis of the real world evidence. Cardiovascular Diabetology. 2019;18(1):80. doi:10.1186/s12933-019-0884-3.
[25] Lin Y, Wang Y, Li P. PPARα: An emerging target of metabolic syndrome, neurodegenerative and cardiovascular diseases. Frontiers in Endocrinology. 2022;13:1074911. doi:10.3389/fendo.2022.1074911.
[26] Kipf TN, Welling M. Semi-Supervised Classification with Graph Convolutional Networks. Proceedings of the International Conference on Learning Representations. 2017.
[27] Veličković P, Cucurull G, Casanova A, Romero A, Liò P, Bengio Y. Graph Attention Networks. Proceedings of the International Conference on Learning Representations. 2018.
[28] Gilmer J, Schoenholz SS, Riley PF, Vinyals O, Dahl GE. Neural Message Passing for Quantum Chemistry. Proceedings of the 34th International Conference on Machine Learning. 2017;70:1263–1272.
[29] Smith Z, Strobel M, Vani BP, Tiwary P. Graph Attention Site Prediction (GrASP): Identifying Druggable Binding Sites Using Graph Neural Networks with Attention. Journal of Chemical Information and Modeling. 2024;64(7):2637–2644. doi:10.1021/acs.jcim.3c01698.
[30] Schütt KT, Sauceda HE, Kindermans P-J, Tkatchenko A, Müller K-R. SchNet – A deep learning architecture for molecules and materials. The Journal of Chemical Physics. 2018;148(24):241722. doi:10.1063/1.5019779.
[31] Gasteiger J, Groß J, Günnemann S. Directional Message Passing for Molecular Graphs. Proceedings of the International Conference on Learning Representations. 2020.
[32] Satorras VG, Hoogeboom E, Welling M. E(n) Equivariant Graph Neural Networks. Proceedings of the 38th International Conference on Machine Learning. 2021;139:9323–9332.
[33] Fuchs F, Worrall D, Fischer V, Welling M. SE(3)-Transformers: 3D Roto-Translation Equivariant Attention Networks. Advances in Neural Information Processing Systems. 2020;33:1970–1981.
[34] Thomas N, Smidt T, Kearnes S, Yang L, Li L, Kohlhoff K, et al. Tensor field networks: Rotation- and translation-equivariant neural networks for 3D point clouds. arXiv. 2018. doi:10.48550/arXiv.1802.08219.
[35] Jing B, Eismann S, Suriana P, Townshend RJL, Dror R. Learning from Protein Structure with Geometric Vector Perceptrons. Proceedings of the International Conference on Learning Representations. 2021.
[36] Kumar R, Romano JD, Ritchie MD. CASTER-DTA: equivariant graph neural networks for predicting drug–target affinity. Briefings in Bioinformatics. 2025;26(5):bbaf554. doi:10.1093/bib/bbaf554.
[37] Stärk H, Ganea O, Pattanaik L, Barzilay DR, Jaakkola T. EquiBind: Geometric Deep Learning for Drug Binding Structure Prediction. Proceedings of the 39th International Conference on Machine Learning. 2022;162:20503–20521.
[38] Powers AS, Yu HH, Suriana P, Koodli RV, Lu T, Paggi JM, et al. Geometric Deep Learning for Structure-Based Ligand Design. ACS Central Science. 2023;9(12):2257–2267. doi:10.1021/acscentsci.3c00572.
[39] Xu M, Yu L, Song Y, Shi C, Ermon S, Tang J. GeoDiff: a Geometric Diffusion Model for

Molecular Conformation Generation. Proceedings of the International Conference on Learning Representations. 2022.
[40] Schneuing A, Harris C, Du Y, Didi K, Jamasb A, Igashov I, et al. Structure-based drug design with equivariant diffusion models. Nature Computational Science. 2024;4(12):899–909. doi:10.1038/s43588-024-00737-x.
[41] Cremer J, Le T, Noé F, Clevert D-A, Schütt KT. PILOT: equivariant diffusion for pocket-conditioned *de novo* ligand generation with multi-objective guidance *via* importance sampling. Chemical Science. 2024;15(36):14954–14967. doi:10.1039/D4SC03523B.
[42] Le T, Cremer J, Clevert D-A, Schütt KT. Equivariant diffusion for structure-based de novo ligand generation with latent-conditioning. Journal of Cheminformatics. 2025;17(1):90. doi:10.1186/s13321-025-01028-x.
[43] Xia Y, Xia C-Q, Pan X, Shen H-B. GraphBind: protein structural context embedded rules learned by hierarchical graph neural networks for recognizing nucleic-acid-binding residues. Nucleic Acids Research. 2021;49(9):e51–e51. doi:10.1093/nar/gkab044.
[44] Gainza P, Sverrisson F, Monti F, Rodolà E, Boscaini D, Bronstein MM, et al. Deciphering interaction fingerprints from protein molecular surfaces using geometric deep learning. Nature Methods. 2020;17(2):184–192. doi:10.1038/s41592-019-0666-6.
[45] Xue F, Zhang M, Li S, Gao X, Wohlschlegel JA, Huang W, et al. SE(3)-equivariant ternary complex prediction towards target protein degradation. Nature Communications. 2025;16(1):5514. doi:10.1038/s41467-025-61272-5.
[46] Li M, Song K, He J, Zhao M, You G, Zhong J, et al. Electron-density-informed effective and reliable de novo molecular design and optimization with ED2Mol. Nature Machine Intelligence. 2025;7(8):1355–1368. doi:10.1038/s42256-025-01095-7.
[47] Aggarwal R, Gupta A, Chelur V, Jawahar CV, Priyakumar UD. DeepPocket: Ligand Binding Site Detection and Segmentation using 3D Convolutional Neural Networks. Journal of Chemical Information and Modeling. 2022;62(21):5069–5079. doi:10.1021/acs.jcim.1c00799.
[48] Marchand A, Buckley S, Schneuing A, Pacesa M, Elia M, Gainza P, et al. Targeting protein–ligand neosurfaces with a generalizable deep learning tool. Nature. 2025;639(8054):522–531. doi:10.1038/s41586-024-08435-4.
[49] Lohmann F, Allenspach S, Atz K, Schiebroek CCG, Hiss JA, Schneider G. Protein Binding Site Representation in Latent Space. Molecular Informatics. 2025;44(1):e202400205. doi:10.1002/minf.202400205.
[50] Yuan Q, Tian C, Song Y, Ou P, Zhu M, Zhao H, et al. GPSFun: geometry-aware protein sequence function predictions with language models. Nucleic Acids Research. 2024;52(W1):W248–W255. doi:10.1093/nar/gkae381.
[51] Wang J, Dokholyan NV. Unified protein–small molecule graph neural networks for binding site prediction. Proceedings of the National Academy of Sciences. 2026;123(10):e2524913123. doi:10.1073/pnas.2524913123.
[52] Kozlovskii I, Popov P. Spatiotemporal identification of druggable binding sites using deep learning. Communications Biology. 2020;3(1):618. doi:10.1038/s42003-020-01350-0.
[53] Lam JH, Katritch V. Navigating structure-based drug discovery with emerging innovations in physics- and knowledge-based approaches. npj Drug Discovery. 2025;2(1):29. doi:10.1038/s44386-025-00031-4.
[54] Özçelik R, Brinkmann H, Criscuolo E, Grisoni F. Generative Deep Learning for de Novo

Drug Design—A Chemical Space Odyssey. Journal of Chemical Information and Modeling. 2025;65(14):7352–7372. doi:10.1021/acs.jcim.5c00641.
[55] Vamathevan J, Clark D, Czodrowski P, Dunham I, Ferran E, Lee G, et al. Applications of machine learning in drug discovery and development. Nature Reviews Drug Discovery. 2019;18(6):463–477. doi:10.1038/s41573-019-0024-5.
[56] Schneider P, Walters WP, Plowright AT, Sieroka N, Listgarten J, Goodnow RA, et al. Rethinking drug design in the artificial intelligence era. Nature Reviews Drug Discovery. 2020;19(5):353–364. doi:10.1038/s41573-019-0050-3.
[57] Wang M, Hsieh C-Y, Wang J, Wang D, Weng G, Shen C, et al. RELATION: A Deep Generative Model for Structure-Based De Novo Drug Design. Journal of Medicinal Chemistry. 2022;65(13):9478–9492. doi:10.1021/acs.jmedchem.2c00732.
[58] Zhang O, Huang Y, Cheng S, Yu M, Zhang X, Lin H, et al. FragGen: towards 3D geometry reliable fragment-based molecular generation. Chemical Science. 2024;15(46):19452–19465. doi:10.1039/D4SC04620J.
[59] Hu Q, Sun C, He H, Xu J, Liu D, Zhang W, et al. Target-aware 3D molecular generation based on guided equivariant diffusion. Nature Communications. 2025;16(1):7928. doi:10.1038/s41467-025-63245-0.
[60] Watson JL, Juergens D, Bennett NR, Trippe BL, Yim J, Eisenach HE, et al. De novo design of protein structure and function with RFdiffusion. Nature. 2023;620(7976):1089–1100. doi:10.1038/s41586-023-06415-8.
[61] Kadan A, Ryczko K, Lloyd E, Roitberg A, Yamazaki T. Guided multi-objective generative AI to enhance structure-based drug design. Chemical Science. 2025;16(29):13196–13210. doi:10.1039/D5SC01778E.
[62] Jin W, Barzilay DR, Jaakkola T. Multi-Objective Molecule Generation using Interpretable Substructures. Proceedings of the 37th International Conference on Machine Learning. 2020;119:4849–4859.
[63] Xie Y, Shi C, Zhou H, Yang Y, Zhang W, Yu Y, et al. MARS: Markov Molecular Sampling for Multi-objective Drug Discovery. Proceedings of the International Conference on Learning Representations. 2021.
[64] Isigkeit L, Hörmann T, Schallmayer E, Scholz K, Lillich FF, Ehrler JHM, et al. Automated design of multi-target ligands by generative deep learning. Nature Communications. 2024;15(1):7946. doi:10.1038/s41467-024-52060-8.
[65] Chen S, Xie J, Ye R, Xu DD, Yang Y. Structure-aware dual-target drug design through collaborative learning of pharmacophore combination and molecular simulation. Chemical Science. 2024;15(27):10366–10380. doi:10.1039/D4SC00094C.
[66] Yuan Y, Pan X, Li X, Zhang R, Su W. A 3D generation framework using diffusion model and reinforcement learning to generate multi-target compounds with desired properties. Journal of Cheminformatics. 2025;17(1):93. doi:10.1186/s13321-025-01035-y.
[67] Zhou X, Guan J, Zhang Y, Peng X, Wang L, Ma J. Reprogramming Pretrained Target-Specific Diffusion Models for Dual-Target Drug Design. Advances in Neural Information Processing Systems. 2024;37:87255–87281. doi: 10.52202/079017-2769.
[68] Shah PM, Zhu H, Lu Z, Wang K, Tang J, Li M. DeepDTAGen: a multitask deep learning framework for drug-target affinity prediction and target-aware drugs generation. Nature Communications. 2025;16(1):5021. doi:10.1038/s41467-025-59917-6.

[69] Zhung W, Kim H, Kim WY. 3D molecular generative framework for interaction-guided drug design. Nature Communications. 2024;15(1):2688. doi:10.1038/s41467-024-47011-2.
[70] Wu J, Qiao A, Wang Z, Wei Z, Chen S. FuseDiff: Symmetry-Preserving Joint Diffusion for Dual-Target Structure-Based Drug Design. arXiv. 2026. doi: 10.48550/arXiv.2603.05567.
[71] Chandak P, Huang K, Zitnik M. Building a knowledge graph to enable precision medicine. Scientific Data. 2023;10(1):67. doi:10.1038/s41597-023-01960-3.
[72] Santos A, Colaço AR, Nielsen AB, Niu L, Strauss M, Geyer PE, et al. A knowledge graph to interpret clinical proteomics data. Nature Biotechnology. 2022;40(5):692–702. doi:10.1038/s41587-021-01145-6.
[73] Jeong C-U, Kim J, Kim D, Sohn K-A. GeOKG: geometry-aware knowledge graph embedding for Gene Ontology and genes. Bioinformatics. 2025;41(4):btaf160. doi:10.1093/bioinformatics/btaf160.
[74] Chen Y, Zhang L. Hi-GeoMVP: a hierarchical geometry-enhanced deep learning model for drug response prediction. Bioinformatics. 2024;40(4):btae204. doi:10.1093/bioinformatics/btae204.
[75] Garnett MJ, Edelman EJ, Heidorn SJ, Greenman CD, Dastur A, Lau KW, et al. Systematic identification of genomic markers of drug sensitivity in cancer cells. Nature. 2012;483(7391):570–575. doi:10.1038/nature11005.
[76] Iorio F, Knijnenburg TA, Vis DJ, Bignell GR, Menden MP, Schubert M, et al. A Landscape of Pharmacogenomic Interactions in Cancer. Cell. 2016;166(3):740–754. doi:10.1016/j.cell.2016.06.017.
[77] Corsello SM, Nagari RT, Spangler RD, Rossen J, Kocak M, Bryan JG, et al. Discovering the anticancer potential of non-oncology drugs by systematic viability profiling. Nature Cancer. 2020;1(2):235–248. doi:10.1038/s43018-019-0018-6.
[78] Barretina J, Caponigro G, Stransky N, Venkatesan K, Margolin AA, Kim S, et al. The Cancer Cell Line Encyclopedia enables predictive modelling of anticancer drug sensitivity. Nature. 2012;483(7391):603–607. doi:10.1038/nature11003.
[79] Su Q, Gou Q, Zhang H, Wang J, Sun H, Hu R, et al. LaMGen: LLM-based 3D molecular generation for multi-target drug design. Nature Communications. 2026;17(1):5091. doi:10.1038/s41467-026-71737-w.
[80] Krishna R, Wang J, Ahern W, Sturmfels P, Venkatesh P, Kalvet I, et al. Generalized biomolecular modeling and design with RoseTTAFold All-Atom. Science. 2024;384(6693):eadl2528. doi:10.1126/science.adl2528.
[81] Lin Z, Akin H, Rao R, Hie B, Zhu Z, Lu W, et al. Evolutionary-scale prediction of atomic-level protein structure with a language model. Science. 2023;379(6637):1123–1130. doi:10.1126/science.ade2574.
[82] Lombard V, Timsit D, Grudinin S, Laine E. SeaMoon: From protein language models to continuous structural heterogeneity. Structure. 2025;33(9):1577-1590.e8. doi:10.1016/j.str.2025.06.010.
[83] Roche R, Tarafder S, Bhattacharya D. Single-sequence protein-RNA complex structure prediction by geometric attention-enabled pairing of biological language models. Cell Systems. 2025;16(10):101400. doi:10.1016/j.cels.2025.101400.
[84] Roche R, Moussad B, Shuvo MH, Tarafder S, Bhattacharya D. EquiPNAS: improved protein–nucleic acid binding site prediction using protein-language-model-informed

equivariant deep graph neural networks. Nucleic Acids Research. 2024;52(5):e27–e27. doi:10.1093/nar/gkae039.
[85] Pang C, Wang Y, Jiang Y, Wang R, Yao X, Zou Q, et al. Multiview Deep Learning-Based Molecule Design and Structural Optimization Accelerates Inhibitor Discover. IEEE Transactions on Neural Networks and Learning Systems. 2025;36(8):14022–14036. doi:10.1109/TNNLS.2024.3506619.
[86] Harding SD, Armstrong JF, Faccenda E, Southan C, Alexander SPH, Davenport AP, et al. The IUPHAR/BPS Guide to PHARMACOLOGY in 2024. Nucleic Acids Research. 2024;52(D1):D1438–D1449. doi:10.1093/nar/gkad944.
[87] Zhou Y, Zhang Y, Zhao D, Yu X, Shen X, Zhou Y, et al. TTD: *Therapeutic Target Database* describing target druggability information. Nucleic Acids Research. 2024;52(D1):D1465–D1477. doi:10.1093/nar/gkad751.
[88] The UniProt Consortium, Bateman A, Martin M-J, Orchard S, Magrane M, Ahmad S, et al. UniProt: the Universal Protein Knowledgebase in 2023. Nucleic Acids Research. 2023;51(D1):D523–D531. doi:10.1093/nar/gkac1052.
[89] Freshour SL, Kiwala S, Cotto KC, Coffman AC, McMichael JF, Song JJ, et al. Integration of the Drug–Gene Interaction Database (DGIdb 4.0) with open crowdsource efforts. Nucleic Acids Research. 2021;49(D1):D1144–D1151. doi:10.1093/nar/gkaa1084.
[90] Szklarczyk D, Kirsch R, Koutrouli M, Nastou K, Mehryary F, Hachilif R, et al. The STRING database in 2023: protein–protein association networks and functional enrichment analyses for any sequenced genome of interest. Nucleic Acids Research. 2023;51(D1):D638–D646. doi:10.1093/nar/gkac1000.
[91] Zhao Y, Xing Y, Zhang Y, Wang Y, Wan M, Yi D, et al. Evidential deep learning-based drug-target interaction prediction. Nature Communications. 2025;16(1):6915. doi:10.1038/s41467-025-62235-6.
[92] Wu K, Xia Y, Deng P, Liu R, Zhang Y, Guo H, et al. TamGen: drug design with target-aware molecule generation through a chemical language model. Nature Communications. 2024;15(1):9360. doi:10.1038/s41467-024-53632-4.
[93] Qian Y, Wu J, Zhang Q. CAT-CPI: Combining CNN and transformer to learn compound image features for predicting compound-protein interactions. Frontiers in Molecular Biosciences. 2022;9:963912. doi:10.3389/fmolb.2022.963912.
[94] Zhavoronkov A, Ivanenkov YA, Aliper A, Veselov MS, Aladinskiy VA, Aladinskaya AV, et al. Deep learning enables rapid identification of potent DDR1 kinase inhibitors. Nature Biotechnology. 2019;37(9):1038–1040. doi:10.1038/s41587-019-0224-x.
[95] Stokes JM, Yang K, Swanson K, Jin W, Cubillos-Ruiz A, Donghia NM, et al. A Deep Learning Approach to Antibiotic Discovery. Cell. 2020;180(4):688-702.e13. doi:10.1016/j.cell.2020.01.021.
[96] Guo AB, Akpinaroglu D, Stephens CA, Grabe M, Smith CA, Kelly MJS, et al. Deep learning–guided design of dynamic proteins. Science. 2025;388(6749):eadr7094. doi:10.1126/science.adr7094.
[97] Nguyen DD, Gao K, Wang M, Wei G-W. MathDL: mathematical deep learning for D3R Grand Challenge 4. Journal of Computer-Aided Molecular Design. 2020;34(2):131–147. doi:10.1007/s10822-019-00237-5.
[98] Zhang J, Durham J, Qian Cong. Revolutionizing protein–protein interaction prediction

with deep learning. Current Opinion in Structural Biology. 2024;85:102775. doi:10.1016/j.sbi.2024.102775.
[99] Liu Z, Wang Y, Vaidya S, Ruehle F, Halverson J, Soljačić M, et al. KAN: Kolmogorov-Arnold Networks. Proceedings of the International Conference on Learning Representations. 2025.
[100] Burley SK, Bhikadiya C, Bi C, Bittrich S, Chen L, Crichlow GV, et al. RCSB Protein Data Bank: powerful new tools for exploring 3D structures of biological macromolecules for basic and applied research and education in fundamental biology, biomedicine, biotechnology, bioengineering and energy sciences. Nucleic Acids Research. 2021;49(D1):D437–D451. doi:10.1093/nar/gkaa1038.
[101] Wang R, Fang X, Lu Y, Wang S. The PDBbind Database: Collection of Binding Affinities for Protein−Ligand Complexes with Known Three-Dimensional Structures. Journal of Medicinal Chemistry. 2004;47(12):2977–2980. doi:10.1021/jm030580l.
[102] Liu H, Zhang W, Zou B, Wang J, Deng Y, Deng L. DrugCombDB: a comprehensive database of drug combinations toward the discovery of combinatorial therapy. Nucleic Acids Research. 2019:gkz1007. doi:10.1093/nar/gkz1007.
[103] Besharatifard M, Vafaee F. A review on graph neural networks for predicting synergistic drug combinations. Artificial Intelligence Review. 2024;57(3):49. doi:10.1007/s10462-023-10669-z.
[104] Wang Y, Wang J, Liu Y. Deep learning for predicting synergistic drug combinations: State-of-the-arts and future directions. Clinical and Translational Discovery. 2024;4(3):e317. doi:10.1002/ctd2.317.
[105] Kim H, Bae B, Park M, Shin Y, Ideker T, Nam H. A genotype-to-drug diffusion model for generation of tailored anti-cancer small molecules. Nature Communications. 2025;16(1):5628. doi:10.1038/s41467-025-60763-9.
[106] Holbeck SL, Camalier R, Crowell JA, Govindharajulu JP, Hollingshead M, Anderson LW, et al. The National Cancer Institute ALMANAC: A Comprehensive Screening Resource for the Detection of Anticancer Drug Pairs with Enhanced Therapeutic Activity. Cancer Research. 2017;77(13):3564–3576. doi:10.1158/0008-5472.CAN-17-0489.
[107] Menden MP, Wang D, Mason MJ, Szalai B, Bulusu KC, Guan Y, et al. Community assessment to advance computational prediction of cancer drug combinations in a pharmacogenomic screen. Nature Communications. 2019;10(1):2674. doi:10.1038/s41467-019-09799-2.
[108] Seo H, Tkachuk D, Ho C, Mammoliti A, Rezaie A, Madani Tonekaboni SA, et al. SYNERGxDB: an integrative pharmacogenomic portal to identify synergistic drug combinations for precision oncology. Nucleic Acids Research. 2020;48(W1):W494–W501. doi:10.1093/nar/gkaa421.
[109] Cai L, Qiao J, Zhou R, Wang X, Li Y, Jiang L, et al. EXPRESSO: a multi-omics database to explore multi-layered 3D genomic organization. Nucleic Acids Research. 2025;53(D1):D79–D90. doi:10.1093/nar/gkae999.
[110] Lv Y, Sun H, Ding Q, Chen B, Ye H, Xu N, et al. Authentication of Linderae Radix through plant metabolomics coupled with a machine learning-enhanced in situ hyperspectral imaging approach. Journal of Pharmaceutical Analysis. 2026;16(5):101476. doi:10.1016/j.jpha.2025.101476.
[111] Huang F, An Y, Zhang L, Wang J, Zhang M, Li Z, et al. UGP system: A deep learning-

driven platform for automated identification of ultrafine granular powders using chromatographic fingerprinting. Journal of Pharmaceutical Analysis. 2026;16(5):101474. doi:10.1016/j.jpha.2025.101474.
[112] Liu Y, Wei Q, Yu G, Gai W, Li Y, Chen X. DCDB 2.0: a major update of the drug combination database. Database. 2014;2014(0):bau124–bau124. doi:10.1093/database/bau124.
[113] Shtar G, Azulay L, Nizri O, Rokach L, Shapira B. CDCDB: A large and continuously updated drug combination database. Scientific Data. 2022;9(1):263. doi:10.1038/s41597-022-01360-z.
[114] Du W, Tang C, Liu G, Sun H, Zhang J, Zhong C. Geometric Structure-Aware Diffusion Model with Self-Optimization Strategy for Molecular Generation. Journal of Chemical Theory and Computation. 2026;22(12):6204–6217. doi:10.1021/acs.jctc.5c02183.
[115] Han T, Wu M, Zhao Q. BIOEMU: AI -Powered Revolution in Scalable Protein Dynamics Simulation. Journal of Cellular and Molecular Medicine. 2025;29(22):e70960. doi:10.1111/jcmm.70960.
[116] Tom G, Schmid SP, Baird SG, Cao Y, Darvish K, Hao H, et al. Self-Driving Laboratories for Chemistry and Materials Science. Chemical Reviews. 2024;124(16):9633–9732. doi:10.1021/acs.chemrev.4c00055.
[117] Li J, Li Y, Chen S, Xing G, Zhang Y, Meng X, et al. Enhancements of symbiotic adhesion and antibiotic efficacy observed by the metabolic crosstalk within cell-bacteria cocultured on a microfluidic gut chip. Journal of Pharmaceutical Analysis. 2026;16(5):101641. doi:10.1016/j.jpha.2026.101641.
[118] Han T, Pan Z, Ge W, Zhao Q. Multimodal Multi-Granularity Fusion Model with Mamba Architecture for Ames Mutagenicity Prediction. Journal of Medicinal Chemistry. 2026;69(2):1766–1778. doi:10.1021/acs.jmedchem.5c03551.
[119] Parrish MC, Tan YJ, Grimes KV, Mochly-Rosen D. Surviving in the Valley of Death: Opportunities and Challenges in Translating Academic Drug Discoveries. Annual Review of Pharmacology and Toxicology. 2019;59(1):405–421. doi:10.1146/annurev-pharmtox-010818-021625.

## Figure Legends

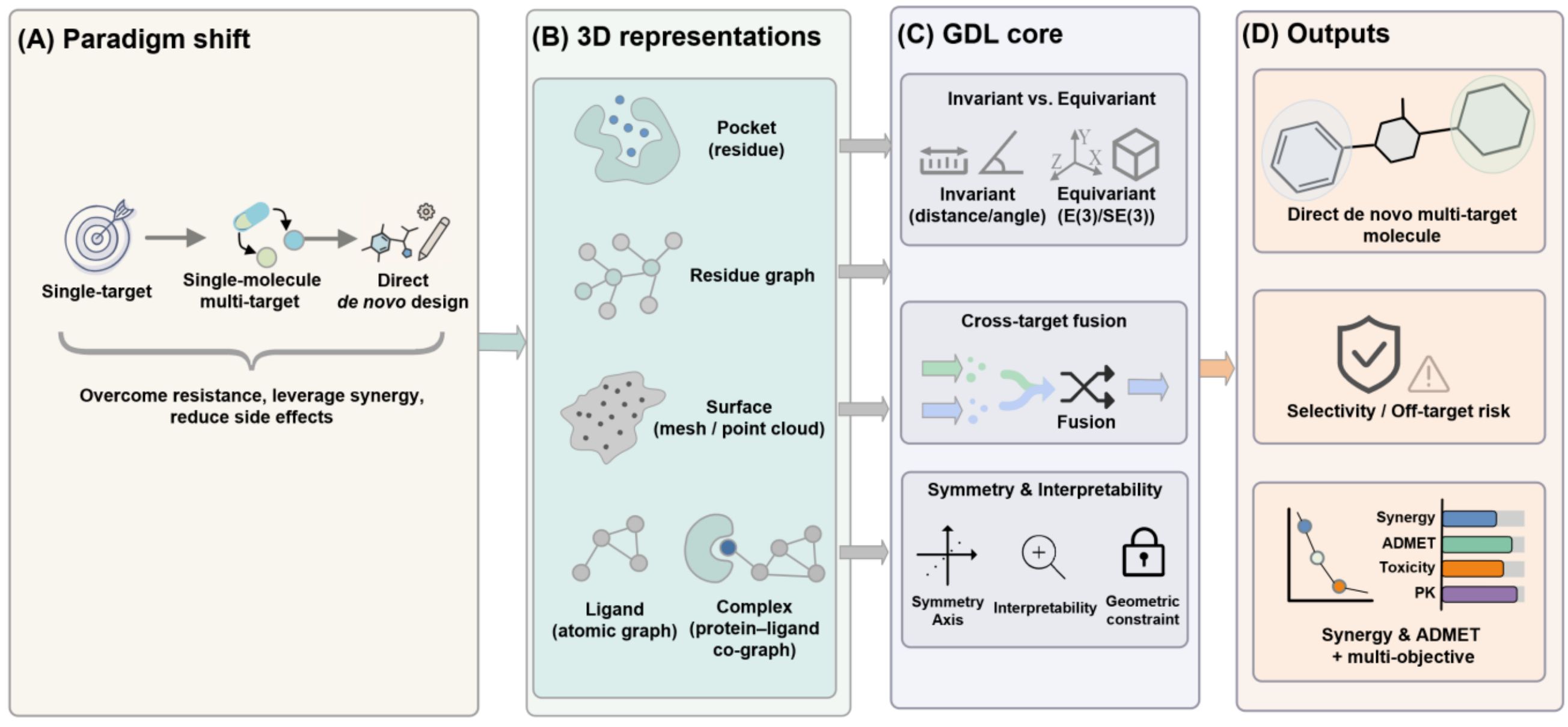

**Figure 1.** Geometric deep learning (GDL) framework enabling the paradigm shift from traditional structure-based drug design (SBDD) to polypharmacology and multi-target drug design. (A) Transition from the conventional "one drug, one target" strategy to single-molecule polypharmacology, and further to the direct de novo generation of multi-target drugs enabled by GDL, thereby overcoming drug resistance, harnessing synergistic effects, and minimizing off-target risks in complex diseases. (B) 3D molecular representations employed in GDL, including residue-level pockets, residue graphs, surface meshes or point clouds, atomic ligand graphs, and protein–ligand co-graphs. (C) Core components of GDL architectures, including invariant versus equivariant modeling (E(3)/SE(3)), cross-target information fusion, and mechanisms for symmetry awareness and interpretability. (D) Primary outputs of GDL-

driven multi-target design, including direct *de novo* generation of multi-target molecules, quantitative assessment of selectivity and off-target risks, and multi-objective optimization integrating synergy, ADMET properties, toxicity, and pharmacokinetic (PK) profiles.

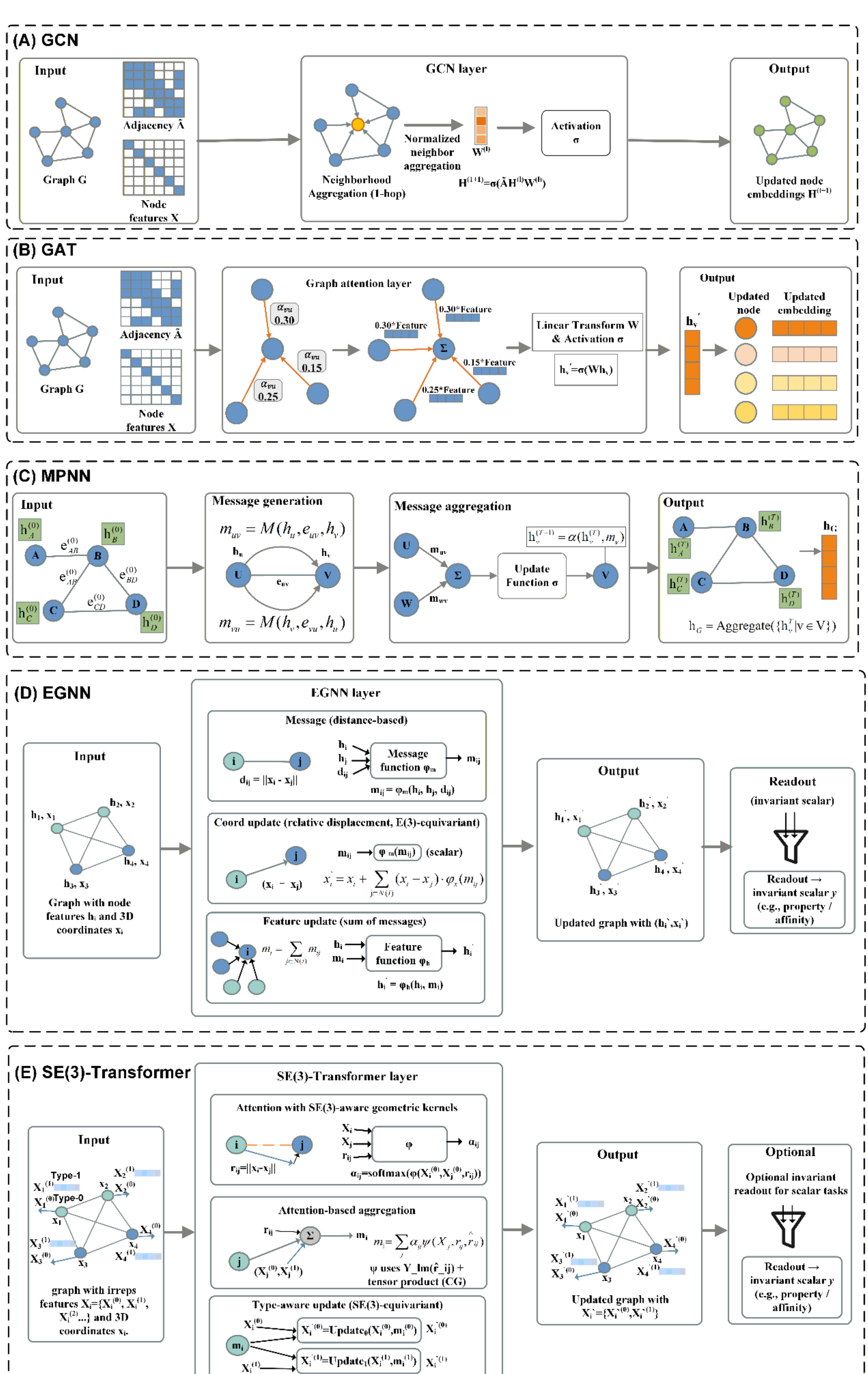

(A) GCN
Input
Adjacency Ã
Graph G
Node features X
GCN layer
Normalized neighbor aggregation
Neighborhood Aggregation (1-hop)
Activation σ
Output
Updated node embeddings
(B) GAT
Input
Adjacency Ã
Graph G
Node features X
Graph attention layer
Linear Transform W & Activation σ
Output
Updated node
Updated embedding
(C) MPNN
Input
Message generation
Message aggregation
Update Function σ
Output
(D) EGNN
EGNN layer
Message (distance-based)
Message function φm
Coord update (relative displacement, E(3)-equivariant)
Feature update (sum of messages)
Feature function φh
Input
Graph with node features hi and 3D coordinates xi
Output
Readout
(invariant scalar)
Readout → invariant scalar y (e.g., property / affinity)
(E) SE(3)-Transformer
SE(3)-Transformer layer
Attention with SE(3)-aware geometric kernels
Attention-based aggregation
ψ uses Y_lm(r̂_ij) + tensor product (CG)
Type-aware update (SE(3)-equivariant)
Input
graph with irreps features and 3D coordinates xi.
Output
Optional
Optional invariant readout for scalar tasks
Readout → invariant scalar y (e.g., property / affinity)

**Figure 2.** Schematic illustrations of representative geometric deep learning architectures in molecular modeling and drug design. (A) Graph Convolutional Network (GCN). (B) Graph Attention Network (GAT). (C) Message Passing Neural Network (MPNN). (D) E(n)-Equivariant Graph Neural Network (EGNN). (E) SE(3)-Transformer. These panels depict the core computational mechanisms of neighborhood aggregation, attention weighting, message passing, equivariant coordinate and feature updates, and SE(3)-aware tensor-product attention, respectively. These architectures form the foundation for subsequent equivariant and generative models discussed in this review.

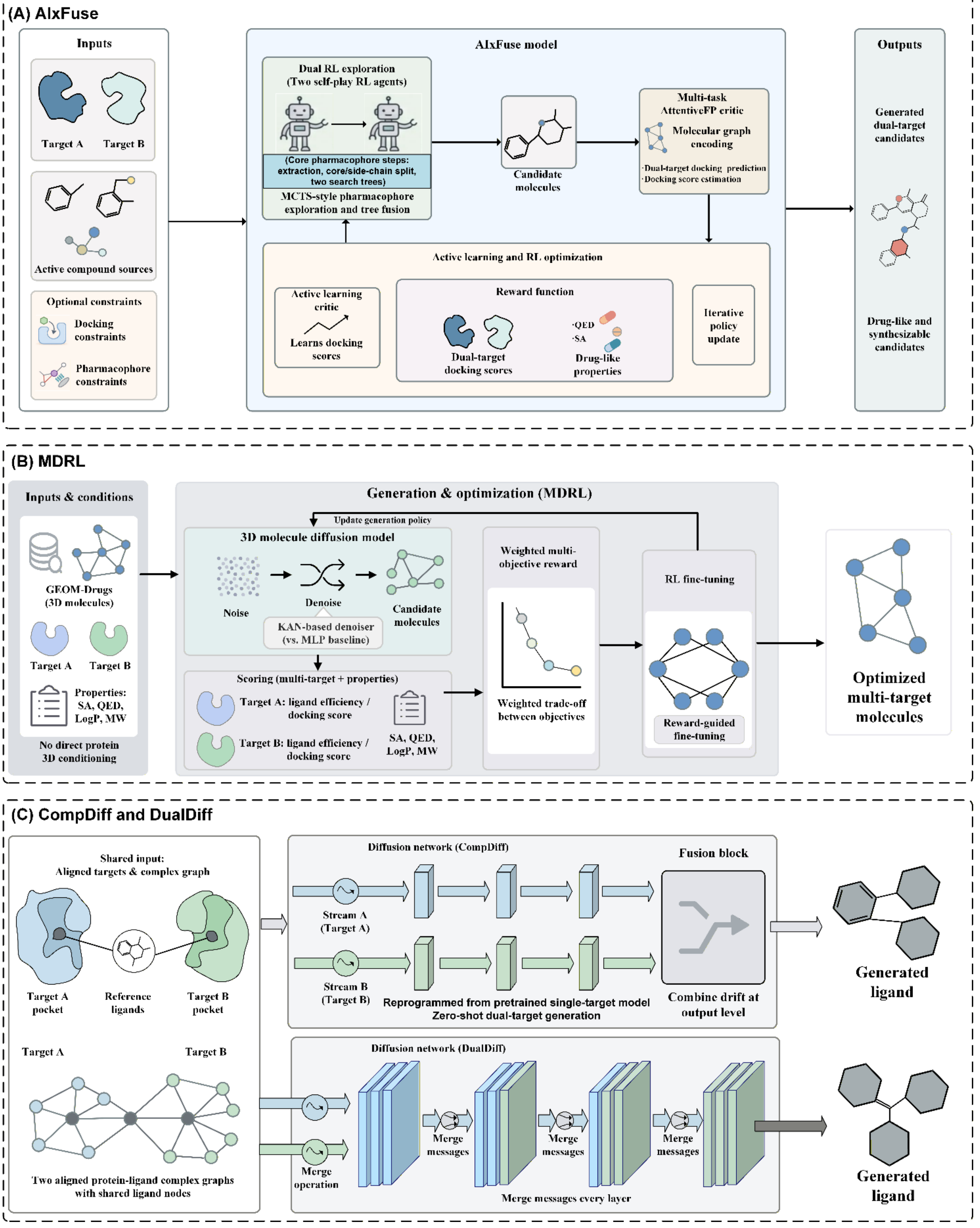

**Figure 3.** Schematic comparison of state-of-the-art structure-aware multi-target ligand generation models using GDL. (A) AIxFuse. External reward

induction via structure-aware dual-target optimization. (B) MDRL. External reward induction via 3D diffusion and reinforcement learning. (C) CompDiff and DualDiff. Internal architectural coupling via equivariant dual-pocket diffusion.

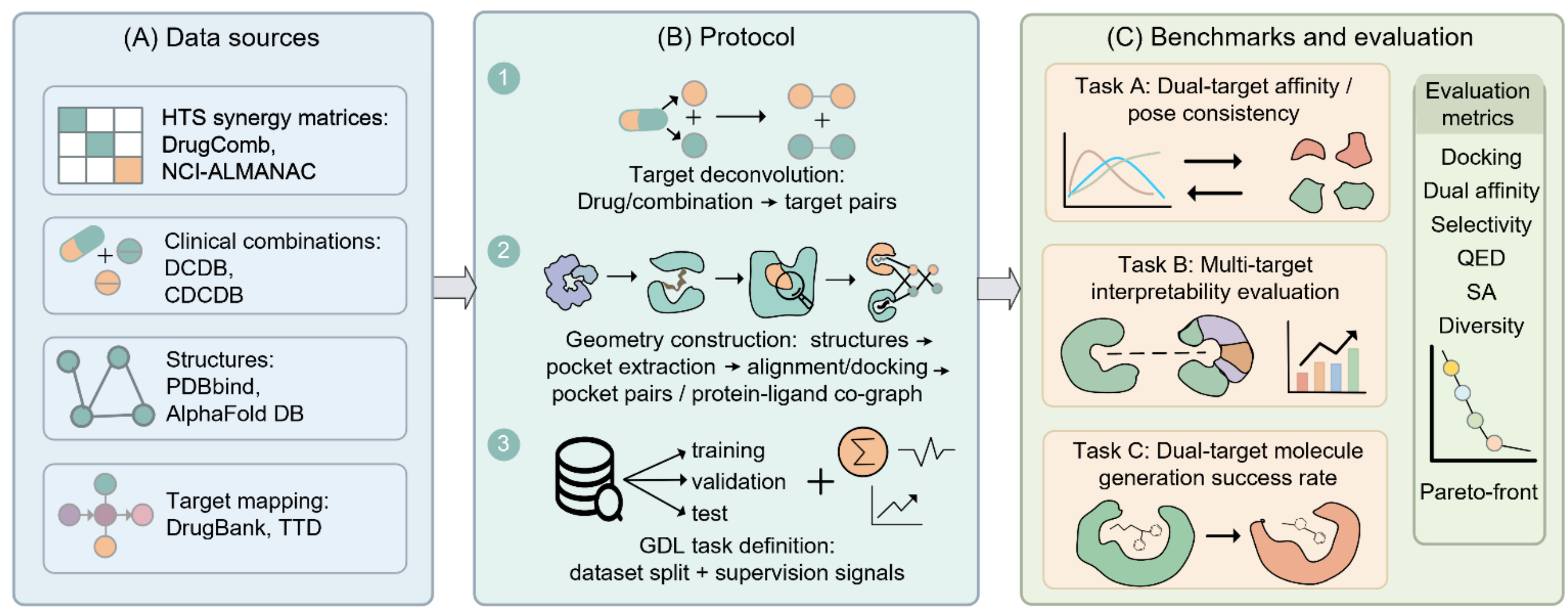

**Figure 4.** Proposed workflow for constructing target–target benchmarks for GDL in polypharmacology and multi-target drug design. (A) Primary data sources comprising high-throughput screening (HTS) synergy matrices (DrugComb, NCI-ALMANAC), clinical combination databases (DCDB 2.0, CDCDB), protein structures (PDBbind, AlphaFold DB), and target mapping resources (DrugBank, TTD). (B) Three-stage protocol: (1) target deconvolution from drug combinations to specific target pairs; (2) geometry construction via pocket extraction, alignment, docking, and formation of pocket-pair or protein–ligand co-graphs; (3) definition of GDL tasks with dataset splitting and synergy-derived supervision signals. (C) Resulting benchmark tasks and evaluation framework, encompassing dual-target affinity and pose consistency prediction, multi-target interpretability evaluation, and dual-target molecule generation success rate, assessed using key metrics such as docking score, dual affinity, selectivity, quantitative estimate of drug-likeness (QED), synthetic accessibility (SA), diversity, and Pareto-front optimization.

## Table Legends

**Table 1.** Representative clinical successes of the "one drug, multiple targets" strategy and their polypharmacological mechanisms.

| **Drug name** | **Primary clinical indications** | **Core target combinations** | **Polypharmacological advantages** |
| --- | --- | --- | --- |
| Clozapine [21] | Treatment-resistant schizophrenia | Dopamine D2/D4 receptors and 5-HT2A receptor | Toxicity reduction and efficacy enhancement: Controls multidimensional psychiatric symptoms while significantly attenuating extrapyramidal side effects commonly associated with single-target agents. |
| Sorafenib [22] | Advanced malignancies (e.g., hepatocellular carcinoma and renal cell carcinoma) | RAS/RAF proliferation pathway and VEGFR/PDGFR receptors | Overcoming drug resistance: Simultaneously inhibits tumor cell proliferation and angiogenesis, thereby blocking compensatory signaling network activation. |
| PPARα/γ dual agonists (e.g., saroglitazar) [24] | Type 2 diabetes mellitus and metabolic syndrome | PPARα (lipid regulation) and PPARγ (insulin sensitization) | Pathway complementarity: Simultaneously improves glycemic control and triglyceride metabolism while avoiding weight gain induced by selective PPARγ agonists. |

**Table 2.** Comparison of representative classic 3D geometric deep learning models in drug design.

| Model | Core idea | Symmetry | Data granularity | Typical applications |
|---|---|---|---|---|
| GCN (2017) [26] | Neighborhood aggregation graph convolution | No built-in invariance (auxiliary features required) | Atomic/residue graphs | Molecular property prediction, binding site detection |
| GAT (2018) [27] | Attention-weighted graph convolution | No built-in invariance | Atomic/residue graphs | Molecular property prediction, binding site detection (e.g., GrASP [29]) |
| MPNN (2017) [28] | General message-passing framework | Distance-invariant (feature-dependent) | Atomic/residue graphs | Quantum property prediction, affinity prediction |
| SchNet (2017) [30] | Continuous convolution with radial basis filters on distances | Rotation and translation invariant (distance-based) | Atomic graphs | Quantum mechanical energy and force prediction, material properties prediction |
| DimeNet (2020) [31] | Directional message passing with three-body angles | Invariant to distances and angles | Atomic graphs | Energy and force prediction |
| EGNN (2021) [32] | Equivariant coordinate and feature updates | E(n)-equivariant | Atomic graphs / point clouds | Molecular conformation generation, protein–ligand modeling |
| SE(3)-Transformer (2020) [33] | Tensor-product attention with spherical harmonics | SE(3)-equivariant | Atomic graphs / point clouds / surface graphs | Protein structure prediction, molecular property prediction |
| GVP-GNN (2020) [35] | Scalar and vector features with rotational equivariance | SE(3)-equivariant | Residue/atomic graphs | Binding-pocket prediction (e.g., PocketMiner [13]), protein design |
| TFN (2018) [34] | Tensor field networks with spherical harmonics convolution | SE(3)-equivariant | Point clouds | Molecular property prediction |
| EquiBind (2022) [37] | Direct equivariant docking pose prediction | SE(3)-equivariant | Atomic graphs (ligand and protein) | Structure-based molecular docking |
| FRAME (2023) [38] | Iterative fragment growth with stepwise equivariant prediction | SE(3)-equivariant (per step) | Atomic graphs (pocket and ligand) | Lead optimization, fragment growing |
| GeoDiff (2022) [39] | Geometric diffusion model for 3D conformations | Approximately SE(3)-invariant (during diffusion) | Point clouds | Molecular conformation sampling and generation |

The “Symmetry” column indicates whether each model is invariant or equivariant. The “Data granularity” column refers to the primary level of molecular representation used.

**Table 3.** Comparison of state-of-the-art structure-aware multi-target ligand generation models.

| Model | Core generative architecture | Multi-target constraint fusion mechanism | Structure and physics perception strategy | Typical advantages and paradigm features |
|---|---|---|---|---|
| AIxFuse (2024) [65] | MCTS exploration and AttentiveFP GNN | External feedback: Pharmacophore-based self-play with multi-task joint scoring | Integration of molecular docking and physical simulations into the generation closed loop | Combines active learning to overcome scarcity of high-quality multi-target data; strong generalization capability |
| DualDiff (2024) [67] | SE(3)-equivariant diffusion model | Internal architecture: Composite graph construction with layer-wise bidirectional equivariant message passing | Rigid-body superposition and alignment of 3D pockets with explicit shared ligand nodes | Zero-shot transfer capability; fundamentally resolves dual-target spatial conflicts at the architectural level |
| MDRL (2025) [66] | KAN-enhanced 3D diffusion model and RL | External reward: Multi-objective reinforcement learning (affinity and drug-likeness metrics) | Docking score-driven strategy fine-tuning; KAN architecture enhances expressiveness and interpretability | Highly flexible; precisely balances multi-target affinity with log *P*, QED and SA properties |
| FuseDiff (2026) [70] | Symmetry-preserving joint diffusion model | Internal architecture: Dual-target local context fusion without global pocket alignment | Dynamic aggregation of geometric contexts from unaligned pockets coupled with explicit chemical bond generation | Alignment-free joint density modeling; strictly enforces topological consistency while generating a shared 2D graph and two independent 3D poses |

**Table 4.** Publicly available drug combination datasets for geometric polypharmacology and multi-target drug design.

| **Dataset name** | **Data scale** | **Dose–response matrix** | **Synergy scores** | **Integrated omics data** | **Typical uses** | **Acquisition** |
|---|---|---|---|---|---|---|
| DrugComb [102] | ~740,000 combinations; 8397 drugs; 2320 cell lines | Yes | Yes (Bliss, HSA, ZIP, Loewe, etc.) | Partial (target annotations and network context) | Synergy prediction, mechanism analysis | Online portal |
| NCI-ALMANAC [106] | 5232 combinations; 104 drugs; 60 cell lines | Yes | Yes (ComboScore) | No (can be supplemented with external NCI-60 data) | Synergy prediction, novel combination discovery | Publication / online repository |
| SYNERGxDB [108] | 22507 combinations; 1977 drugs; 151 cell lines | No | Yes (Bliss, ZIP) | Yes (gene expression, mutations, metabolomics) | Synergy prediction, biomarker discovery | Online portal |
| AstraZeneca–Sanger DREAM Challenge [107] | 910 combinations; 118 drugs; 85 cell lines (~20,000 data points) | Partial (single-agent responses provided) | Yes (-3 to +1) | Yes (gene expression, mutations) | Synergy prediction (competition benchmark); multi-omics analysis | Synapse (registration required) |
| DCDB 2.0 [112] | 1363 combinations; 904 drugs; 814 target pairs | No | No | No | Known combination knowledge; model training | Project website |
| CDCDB [113] | 17107 combinations; 4129 drugs (continuously updated) | No | No | No | Known combination knowledge; real-time synergy prediction | Periodic CSV releases |

The "Dose–response matrix" column indicates whether continuous dose–response data are provided. The "Synergy scores" and "Integrated omics data" columns indicate the presence or absence of these features, with specific metrics or data types detailed in parentheses.